# Understanding Algorithm Performance on an Oversubscribed Scheduling Application

**Laura Barbulescu**                                              LAURABAR@CS.CMU.EDU
*The Robotics Institute*
*Carnegie Mellon University*
*Pittsburgh, PA 15213 USA*
**Adele E. Howe**                                                  HOWE@CS.COLOSTATE.EDU
**L. Darrell Whitley**                                        WHITLEY@CS.COLOSTATE.EDU
**Mark Roberts**                                            MROBERTS@CS.COLOSTATE.EDU
*Computer Science Department*
*Colorado State University*
*Fort Collins, CO 80523 USA*

## Abstract

The best performing algorithms for a particular oversubscribed scheduling application, Air Force Satellite Control Network (AFSCN) scheduling, appear to have little in common. Yet, through careful experimentation and modeling of performance in real problem instances, we can relate characteristics of the best algorithms to characteristics of the application. In particular, we find that plateaus dominate the search spaces (thus favoring algorithms that make larger changes to solutions) and that some randomization in exploration is critical to good performance (due to the lack of gradient information on the plateaus). Based on our explanations of algorithm performance, we develop a new algorithm that combines characteristics of the best performers; the new algorithm's performance is better than the previous best. We show how hypothesis driven experimentation and search modeling can both explain algorithm performance and motivate the design of a new algorithm.

## 1. Introduction

Effective solution of the Air Force Satellite Control Network (AFSCN) oversubscribed scheduling problem runs counter to what works well for similar scheduling problems. Other similar oversubscribed problems, e.g., United States Air Force (USAF) Air Mobility Command (AMC) airlift (Kramer & Smith, 2003) and scheduling telescope observations (Bresina, 1996), are well solved by heuristically guided constructive or repair based search. The best performing solutions to AFSCN are a genetic algorithm (Genitor), Squeaky Wheel Optimization (SWO) and randomized next-descent local search. We have not yet found a constructive or repair based solution that is competitive.

The three best performing solutions to AFSCN appear to have little in common, making it difficult to explain their superior performance. Genitor combines two candidate solutions preserving elements of each. SWO creates an initial greedy solution and then attempts to improve the scheduling of all tasks known to contribute detrimentally to the current evaluation. Randomized local search makes incremental changes based on observed immediate gradients in schedule evaluation. In this paper, we examine the performance of these differ-ent algorithms, identify factors that do or do not help explain the performance and leverage the explanations to design a new search algorithm that is well suited to the characteristics of this application.

Our target application is an oversubscribed scheduling application with alternative resources. AFSCN (Air Force Satellite Control Network) access scheduling requires assigning access requests (communication relays to U.S.A. government satellites) to specific time slots on an antenna at a ground station. It is oversubscribed in that not all tasks can be accommodated given the available resources. To be considered to be oversubscribed, at least some problem instances need to overtax the available resources; for our application though, it appears that most problem instances specify more tasks than can be feasibly scheduled. The application is challenging and shares characteristics with other applications such as Earth Observing Satellites (EOS). It is important in that a team of human schedulers have laboriously performed the task every day for at least 15 years with minimal automated assistance.

All of the algorithms are designed to traverse essentially the same search space: solutions are represented as permutations of tasks, which a greedy schedule builder converts into a schedule by assigning start time and resources to the tasks in the order in which they appear in the permutation. We find that this search space is dominated by large flat regions (plateaus). Additionally, the size of the plateaus increases dramatically as the best solution is approached. The presence of the plateaus indicates that each algorithm needs to effectively manage them in order to find improving solutions.

We have explored a number of different hypotheses to explain the performance of each algorithm. Some of these hypotheses include the following:

**Genitor,** a genetic algorithm, identifies patterns of relative task orderings, similar to backbones from SAT (Singer, Gent, & Smaill, 2000), which are preserved in the members of the population. This is in effect a type of classic building block hypothesis (Goldberg, 1989).

**SWO** starts extremely close to the best solution and so need not enact much change. This hypothesis also implies that it is relatively easy to modify good greedy solutions to find the best known solutions.

**Randomized Local Search** performs essentially a random walk on the plateaus to find exits leading to better solutions; given the distribution of solutions and lack of gradient information, this may be as good a strategy as any.

We tested each of these hypotheses. There is limited evidence for the existence of building blocks or backbone structure. And while Squeaky Wheel Optimization does quickly find good solutions, it cannot reliably find best known solutions. Therefore, while the first two hypotheses were somewhat supported by the data, the hypotheses were not enough to explain the observed performance.

The third hypothesis appears to be the best explanation of why the particular local search strategy we have used works so well. In light of this, we formulated another hypothesis:

**SWO and Genitor** make long leaps in the search space, which allow them to relatively quickly traverse the plateaus.

This last hypothesis appears to well explain the performance of the two methods. For the genetic algorithm the leaps are naturally longer during the early phases of search when parent solutions are less similar.

Based on these studies, we constructed a new search algorithm that exploits what we have learned about the search space and the behavior of successful algorithms. *Attenuated Leap Local Search* makes multiple changes to the solution before evaluating a candidate solution. In addition, the number of changes decreases proportionately with expected proximity to the solution. The number of multiple changes, or the length of the leap, is larger early in the search, and reduces (shortens) as better solutions are found. We find that this algorithm performs quite well: it quickly finds best known solutions to all of the AFSCN problems.

## 2. AFSCN Scheduling

The U.S.A. Air Force Satellite Control Network is currently responsible for coordinating communications between civilian and military organizations and more than 100 USAF managed satellites. Space-ground communications are performed using 16 antennas located at nine tracking stations around the globe[1]. Figure 1 shows a map of the current configuration of AFSCN; this map shows one fewer tracking station and antennae than are in our data, due to those resources apparently having been taken off-line recently. Customer organizations submit task requests to reserve an antenna at a tracking station for a specified time period based on the visibility windows between target satellites and tracking stations. Two types of task requests can be distinguished: low altitude and high altitude orbits. The low altitude tasks specify requests for access to low altitude satellites; such requests tend to be short (e.g., 15 minutes) and have a tight visibility window. High altitude tasks specify requests for high altitude satellites; the durations for these requests are more varied and usually longer, with large visibility windows.

Approximately 500 requests are typically received for a single day. Separate schedules are produced by a staff of human schedulers at Schriever Air Force Base for each day. Of the 500 requests, often about 120 conflicts remain after the first pass of scheduling. "Conflicts" are defined as requests that cannot be scheduled, since they conflict with other scheduled requests (this means that 120 requests remain unscheduled after an initial schedule is produced).

From real problem data, we extract a description of the problem specification in terms of task requests to be scheduled with their corresponding type (low or high altitude), duration, time windows and alternative resources. The AFSCN data also include information about satellite revolution numbers, optional site equipment, tracking station maintenance times (downtimes), possible loss of data due to antenna problems, various comments, etc.; we do not incorporate such information in our problem specification. The information about the type of the task (low or high altitude) as well as the identifier for the satellite involved are included in the task specification. However, we do not know how the satellite identifier corresponds to an actual satellite and so rely on precomputed visibility information which is present in the requests.

---

1. The U.S.A. government is planning to make the AFSCN the core of an Integrated Satellite Control Network for managing satellite assets for other U.S.A. government agencies as well, e.g., NASA, NOAA, other DoD affiliates. By 2011, when the system first becomes operational, the Remote Tracking Stations will be increased and enhanced to accommodate the additional load.

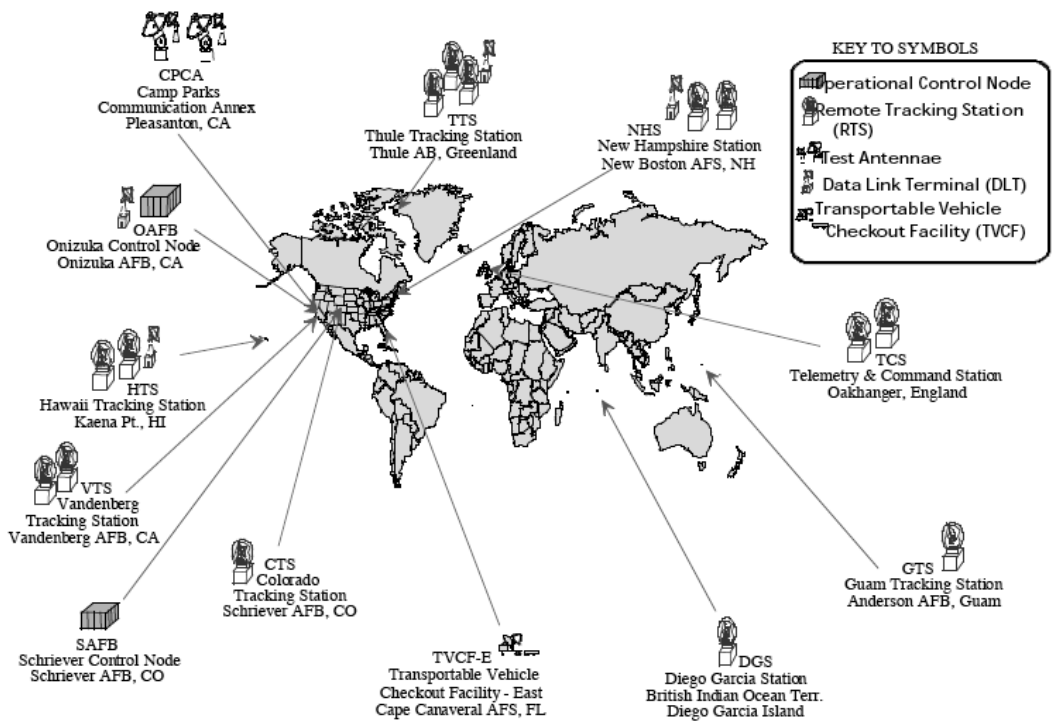

Figure 1: Map of the current AFSCN network including tracking stations, control and relay. The figure was produced for U.S.A. Space and Missile Systems Center (SMC).

A problem instance consists of $n$ task requests. Each task request $T_i$, $1 \leq i \leq n$, specifies a required processing duration $T_i^{Dur}$. Each task request also specifies a number of $j \geq 0$ pairs of the form $(R_j, T_{ij}^{Win})$, each identifying a particular alternative resource (antenna $R_j$) and time window $T_{ij}^{Win}$ for the task. The duration of the task is the same for all possible alternative resources. The start and end of the visibility time window is specific to each alternative resource; therefore while the duration is the same, the time windows can be different for the alternative resources. Once a resource is assigned to the request, the duration needs to be allocated within the corresponding time window. We denote the lower and upper bounds of each time window $j$ corresponding to request $i$ by $T_{ij}^{Win}(LB)$ and $T_{ij}^{Win}(UB)$, respectively. For each task, only one of the alternative antennas needs to be chosen; also, the tasks cannot be preempted once processing is initiated.

While requests are made for a specific antenna, often a different antenna at the same tracking station may serve as an alternate because it has the same capabilities. We assume that all antennas at a tracking station can serve as alternate resources. While this is not always the case in practice, the same assumption was made by previous research from the Air

Force Institute of Technology (AFIT)[2]. A low altitude request specifies as possible resources the antennas present at a single tracking station (for visibility reasons, only one tracking station can accommodate such a request). Usually there are two or three antennas present at a tracking station, and therefore, only two or three possible resources are associated with each of these requests. High altitude requests specify all the antennas present at all the tracking stations that satisfy the visibility constraints; as many as 14 possible alternatives are specified in our data.

Previous research and development on AFSCN scheduling focused on minimizing the number of request conflicts for AFSCN scheduling, or alternatively, maximizing the number of requests that can be scheduled without conflict. Those requests that cannot be scheduled without conflict are bumped out of the schedule. This is not what happens when humans carry out AFSCN scheduling[3]. Satellites are valuable resources, and the AFSCN operators work to fit in every request. What this means in practice is that after negotiation with the customers, some requests are given less time than requested, or shifted to less desirable, but still usable time slots. In effect, the requests are altered until all requests are at least partially satisfied or deferred to another day. By using an evaluation function that minimizes the number of request conflicts, an assumption is being made that we should fit in as many requests as possible before requiring human schedulers to figure out how to place those requests that have been bumped.

However, given that all requests need to be eventually scheduled, we designed a new evaluation criterion that schedules all the requests by allowing them to overlap and minimizing the sum of overlaps between conflicting tasks. This appears to yield schedules that are much closer to those that human schedulers construct. When conflicting tasks are bumped out of the schedule, large and difficult to schedule tasks are most likely to be bumped; placing these requests back into a negotiated schedule means deconstructing the minimal conflict schedule and rebuilding a new schedule. Thus, a schedule that minimizes conflicts may not help all that much when constructing the negotiated schedule, whereas a schedule that minimizes overlaps can suggest ways of fitting tasks into the schedule, for example by reducing a task's duration by two or three minutes, or shifting a start outside of the requested window by a short amount of time.

We obtained 12 days of data for the AFSCN application[4]. The first seven days are from a week in 1992 and were given to us by Colonel James Moore at the Air Force Institute of Technology. These data were used in the first research projects on AFSCN. We obtained an additional five days of data from schedulers at Schriever Air Force Base. Table 2 summarizes the characteristics of the data. The best known solutions were obtained by performing long runs over hundreds of experiments. Using various algorithms and allowing for hundreds of thousands of evaluations, we have not found better solutions than these[5]. We will refer to the problems from 1992 as the $A$ problems, and to the more recent problems, as the $R$ problems.

---

2. In fact, large antennas are needed for high altitude requests, while smaller antennas can handle the low altitude requests. Depending on the type of antennas present at a tracking station, not all antennas can always serve as alternate resources for a request.

3. We met with the several of the schedulers at Schriever to discuss their procedure and have them cross-check our solution. We appreciate the assistance of Brian Bayless and William Szary in setting up the meeting and giving us data.

4. We have approval to make public some, but not all of the data.
   See http://www.cs.colostate.edu/sched/data.html for details on obtaining the problems.

| ID | Date | # Requests | # High | # Low | Best Conflicts | Best Overlaps |
|----|------|-----------|--------|-------|----------------|---------------|
| A1 | 10/12/92 | 322 | 169 | 153 | 8 | 104 |
| A2 | 10/13/92 | 302 | 165 | 137 | 4 | 13 |
| A3 | 10/14/92 | 311 | 165 | 146 | 3 | 28 |
| A4 | 10/15/92 | 318 | 176 | 142 | 2 | 9 |
| A5 | 10/16/92 | 305 | 163 | 142 | 4 | 30 |
| A6 | 10/17/92 | 299 | 155 | 144 | 6 | 45 |
| A7 | 10/18/92 | 297 | 155 | 142 | 6 | 46 |
| R1 | 03/07/02 | 483 | 258 | 225 | 42 | 773 |
| R2 | 03/20/02 | 457 | 263 | 194 | 29 | 486 |
| R3 | 03/26/03 | 426 | 243 | 183 | 17 | 250 |
| R4 | 04/02/03 | 431 | 246 | 185 | 28 | 725 |
| R5 | 05/02/03 | 419 | 241 | 178 | 12 | 146 |

Table 1: Problem characteristics for the 12 days of AFSCN data used in our experiments. ID is used in other tables. Best conflicts and best overlaps are the best known values for each problem for these two objective functions.

## 3. Related Scheduling Research

The AFSCN application is a multiple resource, oversubscribed problem. Examples of other such applications are USAF Air Mobility Command (AMC) airlift scheduling (Kramer & Smith, 2003), NASA's shuttle ground processing (Deale et al., 1994), scheduling telescope observations (Bresina, 1996) and satellite observation scheduling (Frank, Jonsson, Morris, & Smith, 2001; Globus, Crawford, Lohn, & Pryor, 2003).

AMC scheduling assigns delivery missions to air wings (Kramer & Smith, 2003). Their system adopts an iterative repair approach by greedily creating an initial schedule ordering the tasks by priority and then attempting to insert unscheduled tasks by retracting and re-arranging conflicting tasks.

The Gerry scheduler was designed to manage the large set of tasks needed to prepare a space shuttle for its next mission (Zweben, Daun, & Deale, 1994). Tasks are described in terms of resource requirements, temporal constraints and required time windows. The original version used constructive search with dependency-directed backtracking, which was not adequate to the task; a subsequent version employed constraint-directed iterative repair.

In satellite scheduling, customer requests for data collection need to be matched with satellite and tracking station resources. The requests specify the instruments required, the window of time when the request needs to be executed, and the location of the sensing/communication event. These task constraints need to be coordinated with resource constraints; these include the windows of visibility for the satellites, maintenance periods and downtimes for the tracking stations, etc. Typically, more requests need to be scheduled than can be accommodated by the available resources. A general description of the satellite scheduling domain is provided by Jeremy Frank et al. (2001).

---

5. All the best known values can be obtained by running *Genitor* with the population size increased to 400 and allowing 50,000 evaluations per run.

Pemberton (2000) solves a simple one-resource satellite scheduling problem in which the requests have priorities, fixed start times and fixed durations. The objective function maximizes the sum of the priorities of the scheduled requests. A *priority segmentation algorithm* is proposed, which is a hybrid algorithm combining a greedy approach with branch-and-bound. Wolfe and Sorensen (2000) define a more complex one-resource problem, the *window-constrained packing problem* (WCP), which specifies for each request the earliest start time, latest final time and the minimum and maximum duration. The objective function is complex, combining request priority with the position of the scheduled request in its required window and the number of requests scheduled. Two greedy heuristic approaches and a genetic algorithm are implemented; the genetic algorithm is found to perform best.

Globus et al. (2003) compare a genetic algorithm, simulated annealing, Squeaky Wheel Optimization (Joslin & Clements, 1999) and hill climbing on a simplified, synthetic form of the satellite scheduling problem (two satellites with a single instrument) and find that simulated annealing excels and that the genetic algorithm performs relatively poorly. For a general version of satellite scheduling (EOS observation scheduling), Frank et al. (2001) propose a constraint-based planner with a stochastic greedy search algorithm based on Bresina's Heuristic-Biased Stochastic Sampling (HBSS) algorithm (Bresina, 1996). HBSS was originally applied to scheduling astronomy observations for telescopes.

Lemaître et al. (2000) research the problem of scheduling the set of photographs for Agile EOS (ROADEF Challenge, 2003). Task constraints include the minimal time between two successive acquisitions, pairings of requests such that images are acquired twice in different time windows, and hard requirements that certain images must always be acquired. They find that a local search approach performs better than a hybrid algorithm combining branch-and-bound with various domain-specific heuristics.

The AFSCN application was previously studied by researchers from the Air Force Institute of Technology (AFIT). Gooley (1993) and Schalck (1993) described algorithms based on mixed-integer programming (MIP) and insertion heuristics, which achieved good overall performance: $91\% - 95\%$ of all requests scheduled. Parish (1994) used the *Genitor* (Whitley, 1989) genetic algorithm, which scheduled roughly 96% of all task requests, outperforming the MIP approaches. All three of these researchers used the AFIT benchmark suite consisting of seven problem instances, representing actual AFSCN task request data and visibilities for seven consecutive days from October 12 to 18, 1992. Later, Jang (1996) introduced a problem generator employing a bootstrap mechanism to produce additional test problems that are qualitatively similar to the AFIT benchmark problems. Jang then used this generator to analyze the maximum capacity of the AFSCN, as measured by the aggregate number of task requests that can be satisfied in a single-day.

While the general decision problem of AFSCN Scheduling with minimal conflicts is $\mathcal{NP}$-complete, special subclasses of AFSCN Scheduling are polynomial. Burrowbridge (1999) considers a simplified version of AFSCN scheduling, where each task specifies only one resource (antenna) and only low-altitude satellites are present. The objective is to maximize the number of scheduled tasks. Due to the orbital dynamics of low-altitude satellites, the task requests in this problem have negligible *slack*; i.e., the window size is equal to the request duration. Assuming that only one task can be scheduled per time window, the well-known *greedy activity-selector* algorithm (Cormen, Leiserson, & Rivest, 1990) is used to schedule the requests since it yields a solution with the maximal number of scheduled tasks. To schedule low altitude requests on one of the multiple antennas present at a particular ground station, we extended the greedy activity-selector algorithm for multiple resource problems. We proved that this extension of the greedy activity-selector optimally schedules the low altitude requests for the general problem of AFSCN Scheduling (Barbulescu, Watson, Whitley, & Howe, 2004b).

## 4. Algorithms

We implemented a variety of algorithms for AFSCN scheduling: iterative repair, heuristic constructive search, local search, a genetic algorithm (GA), and Squeaky Wheel Optimization (SWO). As will be shown in Section 5, we found that randomized next descent local search, the GA and SWO work best for AFSCN scheduling.

We also considered constructive search algorithms based on texture (Beck, Davenport, Davis, & Fox, 1998) and slack (Smith & Cheng, 1993) constraint-based scheduling heuristics. We implemented straightforward extensions of such algorithms for our application. The results were poor; the number of request tasks combined with the presence of multiple alternative resources for each task make the application of such methods impractical. We do not report the performance values for the constructive search methods because these methods depend critically on the heuristics; we are uncomfortable concluding that the methods are poor because we may not have found good enough heuristics for them. We also tried using a commercial off-the-shelf satellite scheduling package and had similarly poor results. We do not report performance values for the commercial system because it had not been designed specifically for this application and we did not have access to the source to determine the reason for the poor performance.

### 4.1 Solution Representation

Permutation based representations are frequently used when solving scheduling problems (e.g., Whitley, Starkweather, Fuquay, 1989; Syswerda, 1991; Wolfe, Sorensen, 2000; Aickelin, Dowsland, 2003; Globus et al., 2003). All of our algorithms, except iterative-repair, encode solutions using a permutation $\pi$ of the $n$ task request IDs (i.e., [1..n]). A *schedule builder* is used to generate solutions from a permutation of request IDs. The schedule builder considers task requests in the order that they appear in $\pi$. Each task request is assigned to the first resource available from the sequence of resource and window pairs provided in the task description (this is the first feasible resource in the sequence); the earliest possible starting time is then chosen for this resource. When minimizing the number of conflicts, if the request cannot be scheduled on any of the alternative resources, it is dropped from the schedule (i.e., bumped). When minimizing the sum of overlaps, if a request cannot be scheduled without conflict on any of the alternative resources, we assign it to the resource on which the overlap with requests scheduled so far is minimized.[6] Note that our schedule builder does favor the order in which the alternative resources are specified in the request, even though no preference is specified for any of the alternatives.

## 4.2 Iterative Repair

Iterative repair methods have been successfully used to solve various oversubscribed scheduling problems, e.g., Hubble Space Telescope observations (Johnston & Miller, 1994) and space shuttle payloads (Zweben et al., 1994; Rabideau, Chien, Willis, & Mann, 1999). NASA's ASPEN (A Scheduling and Planning Environment) framework (Chien et al., 2000), employs both constructive and repair-based methods and has been used to model and solve real-world space applications such as scheduling EOS. More recently, Kramer and Smith (2003) used repair-based methods to solve the airlift scheduling problem for the USAF Air Mobility Command.

In each case, a key component to the implementation was a domain appropriate ordering heuristic to guide the repairs. For AFSCN scheduling, Gooley's algorithm (1993) uses domain-specific knowledge to implement a repair-based approach. We implement an improvement to Gooley's algorithm that is guaranteed to yield results at least as good as those produced by the original version.

Gooley's algorithm has two phases. In the first phase, the low altitude requests are scheduled, mainly using Mixed Integer Programming (MIP). Because there is a large number of low altitude requests, the requests are divided into two blocks. MIP procedures are first used to schedule the requests in the first block. Then MIP is used to schedule the requests in the second block, which are inserted in the schedule around the requests from the first block. Finally, an interchange procedure attempts to optimize the total number of low altitude requests scheduled. This is needed because the low altitude requests are scheduled in disjoint blocks. Once the low altitude requests are scheduled, their start time and assigned resources remain fixed. In our implementation, we replaced this first phase with a greedy algorithm (Barbulescu et al., 2004b) proven to schedule the optimal number of low altitude requests[7]. Our greedy algorithm modifies the well-known *activity-selector* algorithm (Cormen et al., 1990) for multiple resource problems: the algorithm still schedules the requests in increasing order of their due date, however it specifies that each request is scheduled on the resource for which the idle time before its start time is the minimum. Our version accomplishes the same function as Gooley's first phase, but does so with a guarantee that the optimal number of low-altitude requests are scheduled. Thus, the result is guaranteed to be equal to or better than Gooley's original algorithm.

In the second phase, the high altitude requests are inserted in the schedule (without rescheduling any of the low altitude requests). An order of insertion for the high altitude requests is computed. The requests are sorted in decreasing order of the ratio of the duration of the request to the average length of its time windows (this is similar to the flexibility measure defined by Kramer and Smith, 2003 for AMC); ties are broken based on the number of alternative resources specified (fewer alternatives scheduled first). After all the high altitude requests have been considered for insertion, an interchange procedure attempts to accommodate the unscheduled requests, by rescheduling some of the high altitude requests. For each unscheduled high altitude request, a list of candidate requests for rescheduling is computed (such that after a successful rescheduling operation, the unscheduled request can be placed in the spot initially occupied by such a candidate). A heuristic measure is used to determine which requests from the candidate list should be rescheduled. For the chosen candidates, if no scheduling alternatives are available, the same procedure is applied to identify requests that can be rescheduled. This interchange procedure is defined with two levels of recursion and is called "three satellite interchange".

---

6. If two or more non-scheduled tasks overlap with each other, this mutual overlap is not part of the sum of overlaps. Only the overlap with scheduled requests is considered.

7. Our algorithm optimally solves the problem of scheduling *only* the low altitude requests, in polynomial time.

### 4.3 Randomized Local Search (RLS)

We implemented a hill-climber we call "randomized local search", which starts from a randomly generated solution and iteratively moves toward a better or equally good neighboring solution. Because it has been successfully applied to a number of well-known scheduling problems, we selected a domain-independent move operator, the *shift* operator. From a current solution $\pi$, a neighborhood is defined by considering all $(N-1)^2$ pairs $(x, y)$ of positions in $\pi$, subject to the restriction that $y \neq x - 1$. The neighbor $\pi'$ corresponding to the position pair $(x, y)$ is produced by *shifting* the job at position $x$ into the position $y$, while leaving all other relative job orders unchanged. If $x < y$, then $\pi' = (\pi(1), ..., \pi(x-1), \pi(x+1), ..., \pi(y), \pi(x), \pi(y+1), ..., \pi(n))$. If $x > y$, then $\pi' = (\pi(1), ..., \pi(y-1), \pi(x), \pi(y), ..., \pi(x-1), \pi(x+1), ..., \pi(n))$.

Given the large neighborhood size, we use the shift operator in conjunction with next-descent hill-climbing. Our implementation completely randomizes which neighbor to examine next, and does so with replacement: at each step, both $x$ and $y$ are chosen randomly. This general approach has been termed "stochastic hill-climbing" by Ackley (1987). If the value of the randomly chosen neighbor is equal or better than the value of the current solution, it becomes the new current solution.

It should be emphasized that Randomized Local Search, or stochastic hill-climbing, can sometimes be much more effective than steepest-descent local search or next-descent local search where the neighbors are checked in a predefined order (as opposed to random order). Forrest and Mitchell (1993) showed that a random mutation hill climber (much like our RLS or Ackley's stochastic hill climber) found solutions much faster than steepest-descent local search on a problem they called "The Royal Road" function. The random mutation hill climber also found solutions much faster than a hill climber that generated and examined the neighbors systematically (in a predefined order). Random mutation hill climber was also much more effective than a genetic algorithm for this problem – despite the existence of what would appear to be natural "building blocks" in the function. It is notable that "The Royal Road" function is a staircase like function, where each step in the staircase is a plateau.

### 4.4 Genetic Algorithm

Genetic algorithms were found to perform well on the AFSCN scheduling problem in some early studies (Parish, 1994). Genetic algorithms have also been found to be effective in other oversubscribed scheduling applications, such as scheduling F-14 flight simulators (Syswerda, 1991) or an abstraction of NASA's EOS problem (Wolfe & Sorensen, 2000). For our studies, we used the version of *Genitor* originally developed for a warehouse scheduling application (Starkweather et al., 1991); this is also the version used by Parish for AFSCN scheduling. Like all genetic algorithms, *Genitor* maintains a population of solutions; in our implementation, we fixed the population size to be 200. In each step of the algorithm, a pair of parent solutions is selected, and a crossover operator is used to generate a single child solution, which then replaces the worst solution in the population. Selection of parent solutions is based on the rank of their fitness, relative to other solutions in the population. Following Parish (1994) and Starkweather et al. (1991), we used Syswerda's (1991) position-based crossover operator.

Syswerda's position-based crossover operator starts by selecting a number of random positions in the second parent. The corresponding selected elements will appear in exactly the same positions in the offspring. The remaining positions in the offspring are filled with elements from the first parent in the order in which they appear in this parent:

```
          Parent 1:  A  B  C  D  E  F  G  H  I  J
          Parent 2:  C  F  A  J  H  D  I  G  B  E
Selected Elements:      *  *        *        *
         Offspring:  C  F  A  E  G  D  H  I  B  J
```

For our implementation, we randomly choose the number of positions to be selected, such that it is larger than one third of the total number of positions and smaller than two thirds of the total number of positions.

## 4.5 Squeaky Wheel Optimization

Squeaky Wheel Optimization (SWO) (Joslin & Clements, 1999) repeatedly iterates through a cycle composed of three phases. First, a greedy solution is built, based on priorities associated with the elements in the problem. Then, the solution is analyzed, and the elements causing "trouble" are identified based on their contribution to the objective function. Third, the priorities of such "trouble makers" are modified, such that they will be considered earlier during the next iteration. The cycle is then repeated, until a termination condition is met.

We constructed the initial greedy permutation for SWO by sorting the requests in increasing order of their flexibility. Our flexibility measure is similar to that defined for the AMC application (Kramer & Smith, 2003): the duration of the request divided by the average time window on the possible alternative resources. We break ties based on the number of alternative resources available. For requests with equal flexibilities and numbers of alternative resources, the earlier request is scheduled first. For multiple runs of SWO, we restarted it from a modified permutation created by performing 20 random swaps in the initial greedy permutation.

When minimizing the sum of overlaps, we identified the overlapping requests as the "trouble spots" in the schedule. Note that for any overlap, we considered one request to be scheduled; the other request (or requests, if more than two requests are involved) is "the overlapping request". We sorted the overlapping requests in increasing order of their contribution to the sum of overlaps. We associated with each such request a distance to move forward, based on its rank in the sorted order. We fixed the minimum distance of moving forward to one and the maximum distance to five (this seems to work better than other possible values we tried). The distance values are equally distributed among the ranks. We moved the requests forward in the permutation in increasing order of their contribution to the sum of overlaps: requests with smaller overlaps are moved first. We tried versions of SWO where the distance to move forward is proportional with the contribution to the sum of overlaps or is fixed. However, these versions performed worse than the rank based distance implementation described above. When minimizing conflicts in the schedule all conflicts have an equal contribution to the objective function; therefore we decided to move them forward for a fixed distance of five (we tried values between two and seven but five was best).

## 4.6 Heuristic Biased Stochastic Sampling (HBSS)

HBSS (Bresina, 1996) is an incremental construction algorithm in which multiple root-to-leaf paths are stochastically generated. At each step, the HBSS algorithm needs to heuristically choose the next request to schedule from the unscheduled requests. We used the flexibility measure as described for SWO to rank the unscheduled requests. We compute the flexibility for each request and order them in decreasing order of the flexibility; each request is then given a rank according to this ordering (first request has rank 1, second request rank 2, etc.). A bias function is applied to the ranks; as noted by Bresina (1996, p.271), the choice of bias function "reflects the confidence one has in the heuristic's accuracy - the higher the confidence, the stronger the bias." The flexibility heuristic is an effective greedy heuristic for constructing solutions in AFSCN scheduling. Therefore we used a relatively strong bias function, an exponential bias. For each rank $r$, the bias is computed: $bias(r) = e^{-r}$. The probability to select the unscheduled request with rank $r$ is then computed as:

$$P(r) = \frac{bias(r)}{\sum_{i \in \text{Unscheduled}} bias(rank(i))}$$

where Unscheduled represents the set of unscheduled requests.

Our implementation of HBSS does not re-compute the flexibility of the unscheduled tasks every time we choose the next request to be scheduled. In other words, HBSS is building a permutation of requests and then the schedule builder produces the corresponding schedule. In terms of CPU time, this means that the time required by HBSS to build a solution is similar to those of the other algorithms (dominated by the number of evaluations). A version re-computing the flexibility of the unscheduled tasks as tasks are scheduled would be a lot more expensive. In fact, for EOS which is a similar oversubscribed scheduling problem, Globus et al. (2004) found that updating the heuristic values in HBSS while scheduling was "hundreds of times slower than the permutation-based techniques, required far more memory, and produced very poor schedules".

## 5. What Works Well?

A first step to understanding how best to solve a problem is to assess what methods perform best. The results of running each of the algorithms are summarized in Tables 2 and 3 respectively. For *Genitor*, randomized local search (RLS) and Squeaky Wheel Optimization (SWO), we report the best and mean value and the standard deviation observed over 30 runs, with 8000 evaluations per run. For HBSS, the statistics are taken over 240,000 samples. Both *Genitor* and RLS were initialized from random permutations.

The best known values for the sum of overlaps (see Table 2) were obtained by running *Genitor* with the population size increased to 400 and up to 50,000 evaluations; over hundreds of experiments using numerous algorithms, we have not found better solutions than these. When we report that an algorithm is better than *Genitor* it means that it was better than *Genitor* when both algorithms were limited to 8000 evaluations.

With the exception of Gooley's algorithm, the CPU times are dominated by the number of evaluations and therefore are similar. On a Dell Precision 650 with 3.06 GHz Xeon running Linux, 30 runs with 8000 evaluations per run take between 80 and 190 seconds (for more precise values, see Barbulescu et al., 2004).

The increase in the number of requests received for a day in the more recent $R$ problems causes an increase in the number and percentage of unscheduled requests. For the $A$ problems, at most eight task requests (or 2.5% of the tasks) are not scheduled; between 97.5% and 99% of the task requests are scheduled. For the $R$ problems, at most 42 (or 8.7% of the tasks) are not scheduled; between 91.3% and 97.2% of the tasks requests are scheduled.

To compare algorithm performance, our statistical analyses include *Genitor*, SWO, and RLS. We also include in our analyses the algorithms *SWO1Move* (a variant of SWO we explore in Section 6.5.2), and ALLS (a variant of Local Search we present in Section 7). We judge significant differences of the final evaluations using an ANOVA for the five algorithms on each of the recent days of data. All ANOVAs came back significant, so we are justified in performing pair-wise tests. We examined a single-tailed, two sample t-test as well as the non-parametric Wilcoxon Rank Sum test. The Wilcoxon test significance results were the same as the t-test except in two pairs, so we only present p-values from the t-test that are close to our rejection threshold of $p \leq .005$ *per pair-wise test*[8].

When minimizing conflicts, many of the algorithms find solutions with the best known values. Pair-wise t-tests show that *Genitor* and RLS are not significantly different for R1, R3, and R4. *Genitor* significantly outperforms RLS on R2 ($p = .0023$) and R5 ($p = .0017$). SWO does not perform significantly different from RLS for all five days and significantly outperforms *Genitor* on R5. *Genitor* significantly outperforms SWO on R2 and R4; however, some adjusting of the parameters used to run SWO may fix this problem. It is in fact surprising how well SWO performs when minimizing the conflicts, given that we chose a very simple implementation, where all the tasks in conflict are moved forward with a fixed distance. HBSS performs well for the $A$ problems; however, it fails to find the best known values for R1, R2 and R3. The original solution to the problem, Gooley's, only computes a single solution; its results can be improved by a sampling variant (see Section 6.2.1).

---

8. Five algorithms imply, at worst, 10 pair-wise comparisons per day of data. To control the experiment-wise error, we use a (very conservative, but simple) Bonferroni adjustment; this adjustment is known to increase the probability of a Type II error (favoring false acceptance that the distributions are similar). At $\alpha = .05$, we judge two algorithms as significantly different if $p \leq .005$.

| Day | Genitor | | | RLS | | | SWO | | | HBSS | | | Gooley |
|-----|-----|------|------|-----|-------|------|-----|-------|------|-----|-------|------|--------|
|     | Min | Mean | SD   | Min | Mean  | SD   | Min | Mean  | SD   | Min | Mean  | SD   |        |
| A1  | **8**  | 8.6   | 0.49 | **8**  | 8.7   | 0.46 | **8**  | 8     | 0.0  | **8**  | 9.76  | 0.46 | 11 |
| A2  | **4**  | 4     | 0    | **4**  | 4.0   | 0    | **4**  | 4     | 0.0  | **4**  | 4.64  | 0.66 | 7  |
| A3  | **3**  | 3.03  | 0.18 | **3**  | 3.1   | 0.3  | **3**  | 3     | 0.0  | **3**  | 3.37  | 0.54 | 5  |
| A4  | **2**  | 2.06  | 0.25 | **2**  | 2.2   | 0.48 | **2**  | 2.06  | 0.25 | **2**  | 3.09  | 0.43 | 4  |
| A5  | **4**  | 4.1   | 0.3  | **4**  | 4.7   | 0.46 | **4**  | 4     | 0.0  | **4**  | 4.27  | 0.45 | 5  |
| A6  | **6**  | 6.03  | 0.18 | **6**  | 6.16  | 0.37 | **6**  | 6     | 0.0  | **6**  | 6.39  | 0.49 | 7  |
| A7  | **6**  | 6     | 0    | **6**  | 6.06  | 0.25 | **6**  | 6     | 0.0  | **6**  | 7.35  | 0.54 | **6** |
| R1  | **42** | 43.7  | 0.98 | **42** | 44.0  | 1.25 | 43  | 43.3  | 0.46 | 45  | 48.44 | 1.15 | 45 |
| R2  | **29** | 29.3  | 0.46 | **29** | 29.8  | 0.71 | **29** | 29.96 | 0.18 | 32  | 35.16 | 1.27 | 36 |
| R3  | **17** | 17.63 | 0.49 | **17** | 18.0  | 0.69 | 18  | 18    | 0.0  | 19  | 21.08 | 0.89 | 20 |
| R4  | **28** | 28.03 | 0.18 | **28** | 28.36 | 0.66 | **28** | 28.3  | 0.46 | **28** | 31.22 | 1.10 | 29 |
| R5  | **12** | 12.03 | 0.18 | **12** | 12.4  | 0.56 | **12** | 12    | 0    | **12** | 12.36 | 0.55 | 13 |

Table 2: Performance of *Genitor*, RLS, SWO, HBSS and Gooley's algorithm in terms of the best and mean number of conflicts. Statistics for *Genitor*, local search and SWO are collected over 30 independent runs, with 8000 evaluations per run. For HBSS, 240,000 samples are considered. Min numbers in boldface indicate best known values.

When minimizing overlaps, RLS finds the best known solutions for all but two of the problems. It significantly outperforms *Genitor* on R1 and R2, significantly under-performs on R3, and does not significantly differ in performance on R4 and R5. RLS and SWO do not perform significantly different except for R3 where RLS under-performs. SWO significantly outperforms *Genitor* on all five days. However, if run beyond 8000 evaluations, *Genitor* continues to improve the solution quality but SWO fails to find better solutions. HBSS finds best known solutions to only a few problems. For comparison, we computed the overlaps corresponding to the schedules built using Gooley's algorithm and present them in the last column of Table 3; however, Gooley's algorithm was not designed to minimize overlaps.

## 5.1 Progress Toward the Solution

SWO and Genitor apply different criteria to determine solution modifications. RLS randomly chooses the first shift resulting in an equally good or improving solution. To assess the effect of the differences, we tracked the best value obtained so far when running the algorithms. For each problem, we collected the best value found by SWO, *Genitor* and RLS in increments of 100 evaluations, for 8000 evaluations. We averaged these values over 30 runs of SWO, RLS, and *Genitor*, respectively.

A typical example for each objective function is presented in Figures 2 and 3. For both objective functions, the curves are similar, as is relative performance. SWO quickly finds a good solution, then its performance levels off. RLS progresses quickly during the first half of the search, while *Genitor* exacts smaller improvements. In the second half of the search though, RLS takes longer to find better solutions, while *Genitor* continues to steadily progress toward the best solution. The best so far for Genitor does not improve as quickly as the best so far for RLS. This is not unexpected: the best solution in the Genitor population isn't likely to improve frequently in the beginning of the run. In a sense, tracking the evolution of the median in the population when running Genitor would be more indicative of its progress; we use the best so far to allow for a uniform comparison of the three algorithms.

| Day | Genitor | | | RLS | | | SWO | | | HBSS | | | Gooley |
|-----|-----|------|-----|------|--------|-------|------|-------|------|------|--------|------|------|
| | Min | Mean | SD | Min | Mean | SD | Min | Mean | SD | Min | Mean | SD | |
| A1 | **104** | 106.9 | 0.6 | **104** | 106.76 | 1.81 | **104** | 104 | 0.0 | 128 | 158.7 | 28.7 | 687 |
| A2 | **13** | 13 | 0.0 | **13** | 13.66 | 2.59 | **13** | 13.4 | 2.0 | 43 | 70.1 | 31.1 | 535 |
| A3 | **28** | 28.4 | 1.2 | **28** | 30.7 | 4.31 | **28** | 28.1 | 0.6 | **28** | 52.5 | 16.9 | 217 |
| A4 | **9** | 9.2 | 0.7 | **9** | 10.16 | 2.39 | **9** | 13.3 | 7.8 | **9** | 45.7 | 13.0 | 216 |
| A5 | **30** | 30.4 | 0.5 | **30** | 30.83 | 1.36 | **30** | 30 | 0.0 | 50 | 82.6 | 13.2 | 231 |
| A6 | **45** | 45.1 | 0.4 | **45** | 45.13 | 0.5 | **45** | 45.1 | 0.3 | **45** | 65.5 | 16.8 | 152 |
| A7 | **46** | 46.1 | 0.6 | **46** | 49.96 | 5.95 | **46** | 46 | 0.0 | 83 | 126.4 | 12.5 | 260 |
| R1 | 913 | 987.8 | 40.8 | 798 | 848.66 | 38.42 | 798 | 841.4 | 14.0 | 1105 | 1242.6 | 42.1 | 1713 |
| R2 | 519 | 540.7 | 13.3 | 494 | 521.9 | 20.28 | 491 | 503.8 | 6.5 | 598 | 681.8 | 27.0 | 1047 |
| R3 | 275 | 292.3 | 10.9 | **250** | 327.53 | 55.34 | 265 | 270.1 | 2.8 | 416 | 571.0 | 46.0 | 899 |
| R4 | 738 | 755.4 | 10.3 | **725** | 755.46 | 25.42 | 731 | 736.2 | 3.0 | 827 | 978.4 | 28.7 | 1288 |
| R5 | **146** | 146.5 | 1.9 | **146** | 147.1 | 2.85 | **146** | 146.0 | 0.0 | **146** | 164.4 | 10.8 | 198 |

Table 3: Performance of *Genitor*, local search, SWO, HBSS and Gooley's algorithm in terms of the best and mean sum of overlaps. All statistics are collected over 30 independent runs, with 8000 evaluations per run. For HBSS, 240,000 samples are considered. Min numbers in boldface indicate best known values.

We observe two differences in the objective functions. First, when minimizing the number of conflicts, both *Genitor* and RLS eventually equal or outperform SWO. For minimizing overlaps, *Genitor* and RLS take longer to find good solutions; after 8000 evaluations, SWO found the best solutions. Second, when minimizing the number of conflicts, toward the end of the run, *Genitor* outperforms RLS. When minimizing overlaps, RLS performs better than *Genitor*. Best known solutions for the $R$ problems when minimizing overlaps can be obtained by running RLS for 50,000 evaluations in 30 runs. Running SWO for 50,000 evaluations in 30 runs results in small improvements, and on just two of the problems.

## 6. Hypotheses for Explaining Algorithm Performance

Genitor, SWO and RLS are the most successful algorithms we have tested on the AFSCN problem. Although each operate in the same search space (permutations), they traverse the space rather differently. The puzzle is how can all three be apparently well suited to this problem. To solve the puzzle, first, we describe why plateaus are the dominant feature of the search space. We show that the greedy schedule builder is the main reason for the presence of the plateaus. Then, we test hypotheses that appear to follow from the dominance of the plateaus and the characteristics of each algorithm.

In our study, the greedy schedule builder as well as the objective function are part of the problem specification. Therefore, when formulating and testing our hypotheses, we consider the search space features (such as the plateaus or the number of identical solutions) fixed.

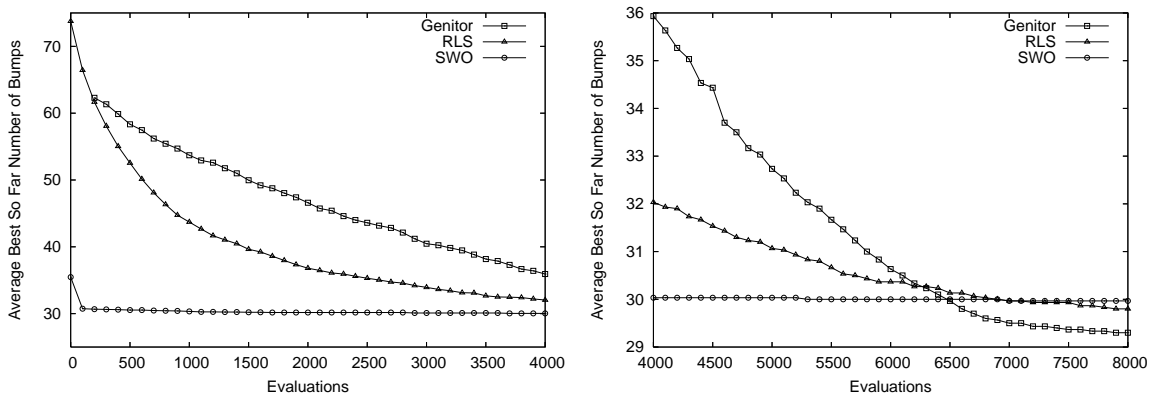

Figure 2: Evolutions of the average best value for conflicts obtained by SWO, RLS and *Genitor* during 8000 evaluations, over 30 runs. The left figure depicts the improvement in the average best value over the first 4000 evaluations. The last 4000 evaluations are depicted in the right figure; note that the scale is different on the y-axis. The curves were obtained for $R2$.

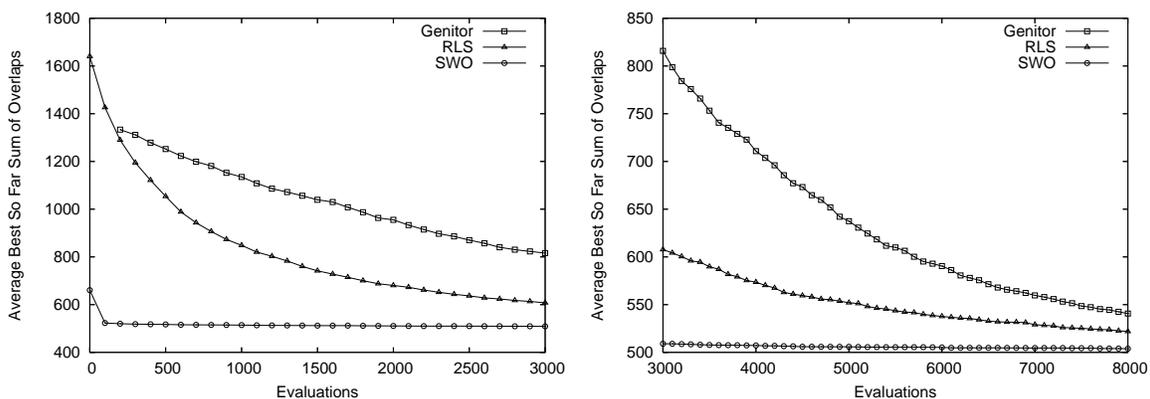

Figure 3: Evolutions of the average best value for sum of overlaps obtained by SWO, RLS and *Genitor* during 8000 evaluations, over 30 runs. The left figure depicts the improvement in the average best value over the first 3000 evaluations. The last 5000 evaluations are depicted in the right figure; note that the scale is different on the y-axis. The curves were obtained for $R2$.

## 6.1 Redundancy of the Search Space

More than a third of the neighbors in RLS result in exactly the same schedule for both the overlaps and minimal conflicts evaluation functions (Barbulescu et al., 2004a; Barbulescu, Whitley, & Howe, 2004c); more than 62% of neighbors in RLS result in the same evaluation (see Section 6.4). The AFSCN search space is dominated by plateaus for three reasons.

The main reason for the presence of plateaus is the greedy schedule builder: each request is scheduled on the first available resource from its list of possible alternatives. For example, consider a permutation of $n-1$ from the total of $n$ requests. If the last request $X$ is inserted in the first position in the permutation and the schedule builder is applied, a schedule $S$ is obtained. We now scan the permutation of $n-1$ requests from left to right, successively inserting $X$ in the second position, then the third and so on, building the corresponding schedule. As long as none of the requests appearing before $X$ in the permutation require the particular spot occupied by $X$ in $S$ as their first feasible alternative to be scheduled, the same schedule $S$ will be obtained. This happens for two reasons: 1) the requests are inserted in the schedule in the order in which they appear in the permutation and 2) the greedy schedule builder considers the possible alternatives in the order in which they are specified and accepts the first alternative for which the request can be scheduled. Let $k+1$ be the first position to insert $X$ that will alter $S$; this means that the first feasible alternative to schedule the request in position $k$ overlaps with the spot occupied by $X$ in $S$. When $X$ is inserted in position $k+1$, a new schedule $S1$ is obtained; the same schedule $S1$ will be built by inserting $X$ in subsequent positions, until encountering a request for which its first feasible alternative overlaps with the spot occupied by $X$ in $S1$, etc. This example also shows that shifting in a permutation might not change the corresponding schedule.

To address the presence of the plateaus in the search space as a result of the greedy schedule builder, we could have used some randomization scheme to diversify the scheduler. However, randomization when implementing a schedule builder can result in problems because of the unpredictability of the value assigned to a permutation. For example, Shaw and Fleming (1997) argue that the use of randomization in a schedule builder can be detrimental to the performance of a genetic algorithm when an indirect representation is used (for which the chromosomes are not schedules, as is the case of *Genitor* for AFSCN scheduling). They support this idea by noting that in general, genetic algorithms rely on the preservation of the good fitness values. Also, for SWO, randomization in the schedule builder changes the significance of reprioritization from one iteration to the next one. If the scheduler is randomized, the new order of requests is very likely to result in a schedule that is not the "repaired version" of the previous one. If the same permutation of requests can be transformed into multiple different schedules because of the nondeterministic nature of the scheduler, the SWO mechanism will not operate as intended.

A second reason for the plateaus in the search space is the presence of time windows. If a request $X$ needs to be scheduled sometime at the end of the day, even if it appears in the beginning of the permutation, it will still occupy a spot in the schedule towards the end (assuming it can be scheduled) and therefore, after most of the other requests (which appeared after $X$ in the permutation).

A third reason is the discretization of the objective function. Clearly, the range of conflicts is a small number of discrete values (with a weak upper bound of the number of tasks). The range for overlaps is still discrete but is larger than for conflicts. Using overlaps as the evaluation function, approximately 20 times more unique objective function values are observed during search compared to searches where the objective is to minimize conflicts. The effect of the discretization can be seen in the differing results using the two objective functions. Thus, one reason for including both in our studies is to show some of the effects of the discretization.

## 6.2 Does Genitor Learn Patterns of Request Ordering?

We hypothesize that *Genitor* performs well because it discovers which interactions between the requests matter. We examine sets of permutations that correspond to schedules with the best known values and identify chains of common request orderings in these permutations, similar in spirit to the notion of backbone in SAT (e.g., Singer et al., 2000). The presence of such chains would support the hypothesis that *Genitor* is discovering patterns of request orderings. This is a classic building block hypothesis: some pattern that is present in parent solutions contributes to their evaluation in some critical way; these patterns are then recombined and inherited during genetic recombination (Goldberg, 1989).

### 6.2.1 COMMON REQUEST ORDERINGS

One of the particular characteristics of the AFSCN scheduling problem is the presence of two categories of requests. The low altitude requests have fixed start times and specify only one to three alternative resources. The high altitude requests implicitly specify multiple possible start times (because their corresponding time windows are usually longer than the duration that needs to be scheduled) and up to 14 possible alternative resources. Clearly the low altitude requests are more constrained. This suggests a possible solution pattern, where low altitude requests would be scheduled first.

To explore the viability of such a pattern, we implemented a heuristic that schedules the low altitude requests before the high altitude ones; we call this heuristic the "split heuristic". We incorporated the split heuristic in the schedule builder: given a permutation of requests, the new schedule builder first schedules only the low altitude requests, in the order in which they appear in the permutation. Without modifying the position of the low altitude requests in the schedule, the high altitude requests are then inserted in the schedule, again in the order in which they appear in the permutation. The idea of scheduling low altitude requests before high altitude requests was the basis of Gooley's heuristic (1993). Also, the split heuristic is similar to the contention measures defined by Frank et al. (2001).

Some of the results we obtained using the split heuristic are surprising: when minimizing conflicts, best known valued schedules can be obtained quickly for the $A$ problems by simply sampling a small number of random permutations. The results obtained by sampling 100 random permutations are shown in Table 4.

While such performance of the split heuristic does not transfer to the $R$ problems or when minimizing the number of overlaps, the results in Table 4 offer some indication of a possible request ordering pattern in good solutions. Is *Genitor* in fact performing well because it discovers that scheduling low before high altitude requests produces good solutions?

As a more general explanation for *Genitor*'s performance, we hypothesize that *Genitor* is discovering patterns of request ordering: certain requests that must come before other requests. To test this, we identify common request orderings present in solutions obtained from multiple runs of *Genitor*. We ran 1000 trials of *Genitor* and selected the solutions corresponding to best known values. First, we checked for request orderings of the form "requestA before requestB" which appear in all the permutations corresponding to best known solutions for the $A$ problems and corresponding to *good solutions* for the $R$ problems. The results are summarized in Table 5. The *Sol. Value* columns show the value of the solutions chosen for the analysis (out of 1000 solutions). The number of solutions (out of 1000) corresponding to the chosen value is shown in the *# of Solutions* columns. When analyzing the common pairs of request orderings for minimizing the number of conflicts, we observed that most pairs specified a low altitude request appearing before a high altitude one. Therefore, we separate the pairs into two categories: pairs specifying a low altitude request before a high altitude requests (column: *(Low,High) Pair Count*) and the rest (column: *Other Pairs*). For the *A* problems, the results clearly show that most common pairs of ordering requests specify a low altitude request before a high altitude request. For the *R* problems, more "Other pairs" can be observed. In part, this might be due to the small number of solutions corresponding to the same value (only 25 out of 1000 for R1 when minimizing conflicts). The small number of solutions corresponding to the same value is also the reason for the big pair counts reported when minimizing overlaps for the *R* problems.

| Day | Best Known | Random Sampling-S | | |
|-----|------------|------|------|-------|
|     |            | Min  | Mean | Stdev |
| A1  | 8          | 8    | 8.2  | 0.41  |
| A2  | 4          | 4    | 4    | 0     |
| A3  | 3          | 3    | 3.3  | 0.46  |
| A4  | 2          | 2    | 2.43 | 0.51  |
| A5  | 4          | 4    | 4.66 | 0.48  |
| A6  | 6          | 6    | 6.5  | 0.51  |
| A7  | 6          | 6    | 6    | 0     |

Table 4: Results of running random sampling with the split heuristic (Random Sampling-S) in 30 experiments, by generating 100 random permutations per experiment for minimizing conflicts.

We know that for the *A* problems the split heuristic results in best-known solutions when minimizing conflicts; therefore, the results in Table 5 are somewhat surprising. We expected to see more low-before-high common pairs of requests for the *A* problems when minimizing the number of conflicts; instead, the pair counts are similar for the two objective functions. *Genitor* seems to discover patterns of request interaction, and most of them specify a low altitude request before a high altitude request.

The results in Table 5 are heavily biased by the number of solutions considered[9]. Indeed, let $s$ denote the number of solutions of identical value (the number in column *# of Solutions*). Also, let $n$ denote the total number of requests. Suppose there are no preferences of orderings between the tasks in good solutions. For a request ordering $A$ before $B$ there is a probability of $1/2$ that it will be present in one of the solutions, and therefore, a probability of $1/2^s$ that it will be present in all $s$ solutions. Given that there exist $n * (n - 1)$ possible precedences, the expected number of common orderings if no preferences of orderings between tasks exist is $n(n - 1)/2^s$. For the A problems and for R5, $s >= 420$. The expected number of common orderings assuming no preferences of orderings between tasks exist is smaller than $n(n-1)/2^{420}$, which is negligible. Therefore, the number of actually detected common precedences (approximately 30 to 125 for low before high pairs and anywhere from 1 to 17 for the others) seem to be actual request patterns. This is also the case for the other R problems. Indeed, for example, for R1, when $s = 15$, the expected number of common orderings if no preferences of orderings between tasks exist is 7.1, while the number of actually detected precedences is 2815 for low before high and 1222 for the other pairs.

---

9. We wish to thank the anonymous reviewer of an earlier version of this work for this insightful observation; the rest of the paragraph is based on his/her comments.

| Day | Minimizing Conflicts | | | | Minimizing Overlaps | | | |
|---|---|---|---|---|---|---|---|---|
| | Sol. Value | # of Solutions | (Low,High) Pair Count | Other Pairs | Sol. Value | # of Solutions | (Low,High) Pair Count | Other Pairs |
| A1 | 8 | 420 | 77 | 1 | 107 | 922 | 78 | 7 |
| A2 | 4 | 1000 | 29 | 1 | 13 | 959 | 50 | 3 |
| A3 | 3 | 936 | 86 | 1 | 28 | 833 | 72 | 10 |
| A4 | 2 | 937 | 132 | 3 | 9 | 912 | 117 | 5 |
| A5 | 4 | 862 | 45 | 9 | 30 | 646 | 48 | 17 |
| A6 | 6 | 967 | 101 | 10 | 45 | 817 | 124 | 10 |
| A7 | 6 | 1000 | 43 | 3 | 46 | 891 | 57 | 11 |
| R1 | 43 | 25 | 2166 | 149 | 947 | 15 | 2815 | 1222 |
| R2 | 29 | 573 | 64 | 5 | 530 | 30 | 1597 | 308 |
| R3 | 17 | 470 | 78 | 21 | 285 | 37 | 1185 | 400 |
| R4 | 28 | 974 | 54 | 16 | 744 | 31 | 1240 | 347 |
| R5 | 12 | 892 | 57 | 10 | 146 | 722 | 109 | 11 |

Table 5: Common pairs of request orderings found in permutations corresponding to best known/good *Genitor* solutions for both objective functions.

The experiment above found evidence to support the hypothesis that *Genitor* solutions exhibit patterns of low before high altitude requests. Given this result, we next investigate if the "split" heuristic (always scheduling low before high altitude requests) can enhance the performance of *Genitor*. To answer this question, we run a second experiment using *Genitor*, where the split heuristic schedule builder is used to evaluate every schedule generated during the search.

Table 6 shows the results of using the split heuristic with *Genitor* on the R problems. *Genitor* with the split heuristic fails to find the best-known solution for *R2* and *R3*. This is not surprising: in fact, we can show that scheduling all the low altitude requests before high altitude requests may prevent finding the optimal solutions.

The results for minimizing sum of overlaps are shown in Table 7. With the exception of A3, A4 and A6, *Genitor* using the split heuristic fails to find best known solutions for the *A* problems. For the *R* problems, using the split heuristic actually improves the results obtained by *Genitor* for R1 and R2; it should be noted that the R1 and R2 solutions are not as good as those found by RLS using 8000 evaluation however. Thus a search that hybridizes the genetic algorithm with a schedule builder using the split heuristic sometimes helps and sometimes hurts in terms of finding good solutions.

We attempted to identify longer chains of common request ordering. We were not successful: while *Genitor* does seem to discover patterns of request ordering, multiple different patterns of request orderings can result in the same conflicts (or even the same schedule).

| Day | Best Known | Genitor with New Schedule Builder | | |
|-----|------------|-----|------|-------|
|     |            | Min | Mean | Stdev |
| R1  | 42         | 42  | 42   | 0     |
| R2  | 29         | 30  | 30   | 0     |
| R3  | 17         | 18  | 18   | 0     |
| R4  | 28         | 28  | 28   | 0     |
| R5  | 12         | 12  | 12   | 0     |

Table 6: Minimizing conflicts: results of running *Genitor* with the split heuristic in 30 trials, with 8000 evaluations per trial.

| Day | Best Known | Genitor with New Schedule Builder | | |
|-----|------------|-----|--------|-------|
|     |            | Min | Mean   | Stdev |
| A1  | 104        | 119 | 119    | 0.0   |
| A2  | 13         | 43  | 43     | 0.0   |
| A3  | 28         | 28  | 28     | 0.0   |
| A4  | 9          | 9   | 9      | 0.0   |
| A5  | 30         | 50  | 50     | 0.0   |
| A6  | 45         | 45  | 45     | 0.0   |
| A7  | 46         | 69  | 69     | 0.0   |
| R1  | 774        | 907 | 924.33 | 6.01  |
| R2  | 486        | 513 | 516.63 | 5.03  |
| R3  | 250        | 276 | 276.03 | 0.18  |
| R4  | 725        | 752 | 752.03 | 0.0   |
| R5  | 146        | 146 | 146    | 0.0   |

Table 7: Minimizing sum of overlaps: results of running *Genitor* with the split heuristic using the split heuristic schedule builder to evaluate each schedule. The results are based on 30 experiments, with 8000 evaluations per experiment.

We could think of these patterns as building blocks. *Genitor* identifies good building blocks (orderings of requests resulting in good partial solutions) and propagates them into the final population (and the final solution). Such patterns are essential in building a good solution. However, the patterns are not ubiquitous (not all of them are necessary) and, therefore, attempts to identify them across different solutions produced by *Genitor* have failed.

| Day | Minimizing Conflicts | | | | Minimizing Overlaps | | | |
|---|---|---|---|---|---|---|---|---|
| | Best Known | Min | Mean | Stdev | Best Known | Min | Mean | Stdev |
| A1 | 8 | 8 | 8.0 | 0.0 | 104 | 104 | 104.46 | 0.68 |
| A2 | 4 | 4 | 4.0 | 0.0 | 13 | 13 | 13.83 | 1.89 |
| A3 | 3 | 3 | 3.16 | 0.46 | 28 | 28 | 30.13 | 1.96 |
| A4 | 2 | 2 | 2.13 | 0.34 | 9 | 9 | 11.66 | 1.39 |
| A5 | 4 | 4 | 4.03 | 0.18 | 30 | 30 | 30.33 | 0.54 |
| A6 | 6 | 6 | 6.23 | 0.63 | 45 | 45 | 48.3 | 6.63 |
| A7 | 6 | 6 | 6.0 | 0.0 | 46 | 46 | 46.26 | 0.45 |
| R1 | 42 | 42* | 43.43 | 0.56 | 774 | 851 | 889.96 | 31.34 |
| R2 | 29 | 30 | 30.1 | 0.3 | 486 | 503 | 522.2 | 9.8 |
| R3 | 17 | 17* | 17.73 | 0.44 | 250 | 268 | 276.4 | 4.19 |
| R4 | 28 | 28 | 28.53 | 0.57 | 725 | 738 | 758.26 | 12.27 |
| R5 | 12 | 12 | 13.1 | 0.4 | 146 | 147 | 151.03 | 2.19 |

Table 8: Statistics for the results obtained in 30 runs of SWO initialized with random permutations (i.e., *RandomStartSWO*), with 8000 evaluations per run. The mean and best value from 30 runs as well as the standard deviations are shown. The entries with a * indicate values that are better than the corresponding SWO values. For each problem, the best known solution for each objective function is also included.

### 6.3 Is SWO's Performance Due to Initialization?

The graphs of search progress for SWO (Figures 2 and 3) show that it starts with much better solutions than do the other algorithms. The initial greedy solution for SWO translated into best known values for five problems (A2, A3, A5, A6 and R5) when minimizing the number of conflicts and for two problems (A6 and R5) when minimizing overlaps.

How important is the initial greedy permutation for SWO? To answer this question, we replaced the initial greedy permutation (and its variations in subsequent iterations of SWO) with random permutations and then used the SWO mechanism to iteratively move forward the requests in conflict. We call this version of SWO *RandomStartSWO*. We compared the results produced by *RandomStartSWO* with results from SWO to assess the effects of the initial greedy solution. The results produced by *RandomStartSWO* are presented in Table 8. The entries with a * indicate that *RandomStartSWO* produced a better result than SWO. With the exception of R2, when minimizing the number of conflicts, best known values are obtained by *RandomStartSWO* for all the problems. In fact, for R1 and R3, the best results obtained are slightly better than the best found by SWO. When minimizing the sum of overlaps, best known values are obtained for the *A* problems; only for the *R* problems, the performance of SWO worsens when it is initialized with a random permutation. However, *RandomStartSWO* still performs better or as well as *Genitor* (with the exception of R2 when minimizing the number of conflicts and R5 for overlaps) for both objective functions. These results suggest that the initial greedy permutation is not the main performance factor for SWO: the performance of *RandomStartSWO* is competitive with that of *Genitor*.

| Day | Total Neighbors | Minimizing Conflicts | | | | Minimizing Overlaps | | | |
|-----|-----------------|----------------------|--------|----------------------|--------|---------------------|--------|--------------|--------|
|     |                 | Random Perms | | Optimal Perms | | Random Perms | | Optimal Perms | |
|     |                 | Mean | Avg % | Mean | Avg % | Mean | Avg % | Mean | Avg % |
| A1 | 103041 | 87581.1 | 84.9 | 91609.1 | 88.9 | 75877.4 | 73.6 | 88621.2 | 86.0 |
| A2 | 90601 | 79189.3 | 87.4 | 83717.9 | 92.4 | 70440.9 | 77.7 | 81141.9 | 89.5 |
| A3 | 96100 | 82937 | 86.8 | 84915.4 | 88.9 | 73073.3 | 76.5 | 82407.7 | 86.3 |
| A4 | 100489 | 84759 | 84.3 | 87568.2 | 87.1 | 72767.7 | 72.4 | 85290 | 84.8 |
| A5 | 92416 | 77952 | 84.3 | 82057.4 | 88.7 | 67649.3 | 73.2 | 79735.9 | 86.2 |
| A6 | 88804 | 74671.5 | 84.0 | 78730.3 | 88.6 | 63667.4 | 71.6 | 75737.9 | 85.2 |
| A7 | 87616 | 76489.6 | 87.3 | 79756.5 | 91.0 | 67839 | 77.4 | 77584.3 | 88.5 |
| R1 | 232324 | 189566 | 81.5 | 190736 | 82.0 | 145514 | 62.6 | 160489 | 69.0 |
| R2 | 207936 | 173434 | 83.4 | 177264 | 85.2 | 137568 | 66.1 | 160350 | 77.1 |
| R3 | 180625 | 153207 | 84.8 | 156413 | 86.5 | 126511 | 70.0 | 139012 | 76.9 |
| R4 | 184900 | 157459 | 85.1 | 162996 | 88.1 | 130684 | 70.6 | 145953 | 78.9 |
| R5 | 174724 | 154347 | 88.3 | 159581 | 91.3 | 133672 | 76.5 | 152629 | 87.3 |

Table 9: Statistics for the number of neighbors resulting in schedules of the same value as the original, over 30 random and optimal permutations, for both objective functions

## 6.4 Is RLS Performing a Random Walk?

RLS spends most of the time traversing plateaus in the search space (by accepting non-improving moves). In this section, we study the average length of random walks on the plateaus encountered by local search. We show that as search progresses the random walks become longer before finding an improvement, mirroring the progress of RLS. We note that a similar phenomenon has been observed for SAT (Frank, Cheeseman, & Stutz, 1997).

More than a third of all shifting pairs of requests result in schedules identical with the current solution (Barbulescu et al., 2004a, 2004c). However, an even larger number of neighbors result in different schedules with the same value as the current solution. This means that most of the accepted moves during search are non-improving moves; search ends up randomly walking on a plateau until an exit is found. We collected results about the number of schedules with the same *value* as the original schedule, when perturbing the solutions in all possible pairwise changes. Note that these schedules include the ones identical with the current solution. The results are summarized in Table 9. We report the average percentage of neighbors identical in value with the original permutation. The results show that: 1) More than 84% of the shifts result in schedules with the same value as the original one, when minimizing conflicts. When minimizing overlaps, more than 62% (usually around 70%) of the shifts result in same value schedules. 2) Best known solutions have slightly more same-value neighbors than do random permutations; the difference is statistically significant when minimizing overlaps. This suggests that the plateaus corresponding to good values in the search space might be larger in size than the plateaus corresponding to random permutations.

To assess the size of the plateaus and their impact on RLS, we performed random walks at fixed intervals during RLS. At every 500 evaluations of RLS, we identified the current solution $Crt$. For each such $Crt$, we performed 100 iterations of local search starting from $Crt$ and stopping as soon as a better solution or a maximum number of equally good solutions were encountered. For the $A$ problems, best known solutions are often found early in the search; most of the 100 iterations of local search started from such a $Crt$ would reach the maximum number of equally good solutions. Therefore, we chose a limit of 1000 steps on the plateau for the $A$ problems and 8000 steps for the $R$ problems. We averaged the number of equally good solutions encountered during the 100 trials of search performed for each $Crt$; this represents the average number of steps needed to find an exit from a plateau.

Figure 4 displays the results obtained for R4; similar behavior was observed for the rest of the problems. Note that we used a $log$ scale on the $y$ axis for the graph corresponding to minimizing overlaps: most of the 100 walks performed from the current solution of value 729 end up taking the maximum number of steps allowed (8000) without finding an exit from the plateau. Also, the random walk steps counts only equal moves; the number of evaluations needed by RLS (x-axis) is considerably higher due to needing to check detrimental moves before accepting equal ones. The results show that large plateaus are present in the search space; improving moves lead to longer walks on lower plateaus, which when detrimental moves are factored in, appears to mirror the performance of RLS.

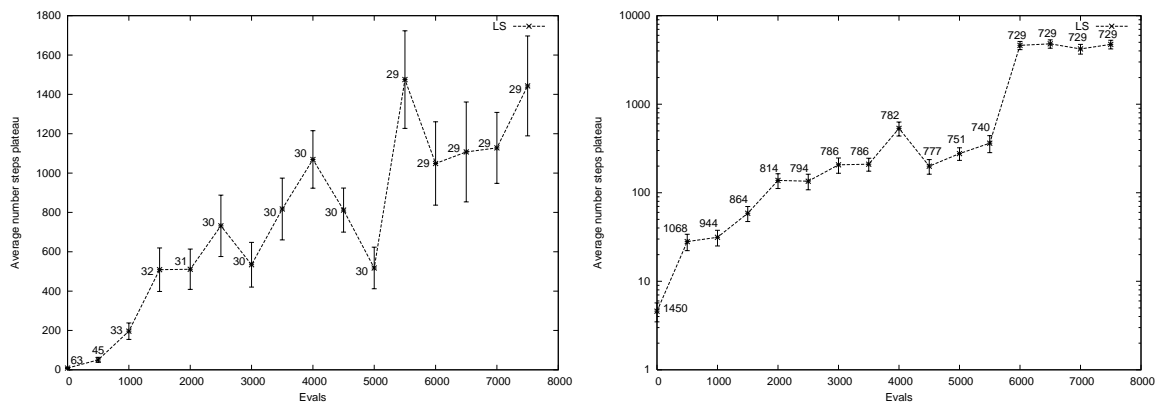

Figure 4: Average length of the random walk on plateaus when minimizing conflicts (left) or overlaps (right) for a single local search run on R4. The labels on the graphs represent the value of the current solution. Note the $log$ scale on the $y$ axis for the graph corresponding to minimizing overlaps. The best known value for this problem is 28 when minimizing conflicts and 725 when minimizing overlaps.

For the AFSCN scheduling problems, most of the states on a plateau have at least one neighbor that has a better value (this neighbor represents an exit). However, the number of such exits is a very small percentage of the total number of neighbors, and therefore, local search has a very small probability of finding an exit. Using the terminology introduced by Frank et al. (1997), most of the plateaus encountered by search in the AFSCN domain would be classified as benches, meaning that exits to states at lower levels are present. If there are no exits from a plateau, the plateau is a local minimum. Determining which of the plateaus are local minima (by enumerating all the states on the plateau and their neighbors) is prohibitive because of the large size of the neighborhoods and the large number of equally good neighbors present for each state in the search space. Instead, we focus on the average length of the random walk on a plateau as a factor in local search performance. The length of the random walk on the plateau depends on two features: the size of the plateau and the number of exits from the plateau. Preliminary investigations show that the number of improving neighbors for a solution decreases as the solution becomes better - therefore we conjecture that there are more exits from higher level plateaus than from the lower level ones. This would account for the trend of needing more steps to find an exit when moving to lower plateaus (corresponding to better solutions). It is also possible that the plateaus corresponding to better solutions are larger in size; however, enumerating all the states on a plateau for the AFSCN domain is impractical (following a technique developed by Frank et al., 1997, just the first iteration of breadth first search would result in approximately $0.8 * (n-1)^2$ states on the same plateau).

## 6.5 Are Long Leaps Instrumental?

As in other problems with large plateaus (e.g., research published by Gent and Walsh, 1995 on SAT), we hypothesize that long leaps in the search space are instrumental for an algorithm to perform well on AFSCN scheduling. SWO is moving forward multiple requests that are known to be problematic. The position crossover mechanism in *Genitor* can be viewed as applying *multiple* consecutive shifts to the first parent, such that the requests in the selected positions of the second parent are moved into those selected positions of the first. In a sense, each time the crossover operator is applied, a multiple move is proposed for the first parent. We hypothesize that this multiple move mechanism present in both SWO and *Genitor* allows them to make long leaps in the space and thus reach solutions fast.

Note that if we knew exactly which requests to move, moving forward only a small number of requests (or even only one) might be all that is needed to reach the solutions quickly. Finding which requests to move is difficult; in fact we studied the performance of a more informed move operator that only moves requests into positions which guarantee schedule changes (Roberts et al., 2005). We found surprising results: the more informed move operator performs worse than the random unrestricted shift employed by RLS. We argue that the multiple moves are a desired algorithm feature as they make it more likely that one of the moves will be the *right* one.

To investigate our hypothesis about the role of multiple moves when traversing the search space, we perform experiments with a variable number of moves at each step for both *Genitor* and SWO. For *Genitor*, we vary the number of crossover positions allowed. For SWO, we vary the number of requests in conflict moved forward.

### 6.5.1 The Effect of Multiple Moves on *Genitor*

To test the effect of multiple moves on *Genitor*, we change Syswerda's position crossover by imposing a fixed number of selected positions in the second parent (see Section 4.4 for a description of Syswerda's position crossover). We call this implementation *Genitor-k* where k is the number of selected positions. Recall that our implementation of Syswerda's position crossover randomly selects a number of positions that is larger than one third and smaller than two thirds of the total number of positions. If multiple moves are indeed a factor in performance then increasing the number of selected positions up to a point should result in finding improvements faster. If only a few positions are selected, the offspring will be very similar to the first parent. If the number of selected positions is large, close to the number of total requests, the offspring will be very similar to the second parent. If the offspring is very similar to one of its two parents, we expect a slower rate in finding improvements to the current best solution. Therefore, both for small and for large *k* values, we expect *Genitor-k* to find improvements at a much slower rate than *Genitor* or *Genitor-k* with average *k* values (values closer to half of the number of requests).

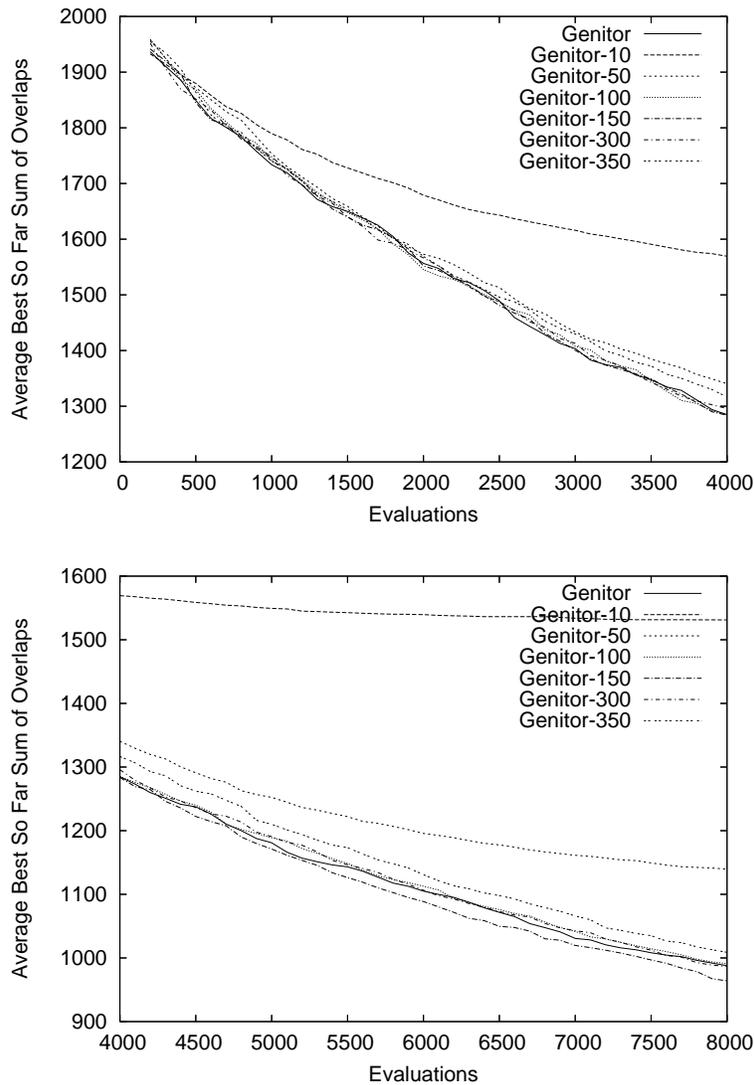

Figure 5: Evolutions of the average best value obtained by *Genitor* and its versions with a fixed number of selected positions for crossover. during 8000 evaluations, over 30 runs. The graphs were obtained for *R*1; best solution value is 773.

| Day | Genitor-10 | | | Genitor-50 | | | Genitor-100 | | | Genitor-150 | | |
|-----|-----|------|-------|-----|-------|-------|-----|-------|-------|-----|-------|-------|
| | Min | Mean | Stdev | Min | Mean | Stdev | Min | Mean | Stdev | Min | Mean | Stdev |
| A1 | 11 | 14.93 | 1.94 | **8** | 9.26 | 0.63 | **8** | 8.66 | 0.47 | **8** | 8.53 | 0.5 |
| A2 | 5 | 7.13 | 1.77 | **4** | 4.03 | 0.18 | **4** | 4.0 | 0.0 | **4** | 4.0 | 0.0 |
| A3 | 6 | 10.4 | 2.12 | **3** | 3.36 | 0.55 | **3** | 3.0 | 0.0 | **3** | 3.0 | 0.0 |
| A4 | 5 | 10.66 | 2.7 | **2** | 3.13 | 0.81 | **2** | 2.23 | 0.50 | **2** | 2.06 | 0.25 |
| A5 | 5 | 9.6 | 2.29 | **4** | 4.73 | 0.69 | **4** | 4.26 | 0.44 | **4** | 4.2 | 0.4 |
| A6 | 9 | 12.63 | 1.8 | **6** | 6.83 | 0.94 | **6** | 6.03 | 0.18 | **6** | 6.06 | 0.25 |
| A7 | 8 | 10.6 | 1.75 | **6** | 6.1 | 0.30 | **6** | 6.0 | 0.0 | **6** | 6.0 | 0.0 |
| R1 | 57 | 66.5 | 4.38 | 47 | 52.0 | 2.82 | **42** | 45.83 | 1.68 | **42** | 44.36 | 1.24 |
| R2 | 42 | 47.16 | 3.59 | 32 | 34.53 | 1.47 | **29** | 30.0 | 0.78 | **29** | 29.6 | 0.56 |
| R3 | 27 | 31.1 | 2.41 | 19 | 21.6 | 1.67 | **17** | 18.03 | 0.61 | **17** | 17.63 | 0.61 |
| R4 | 36 | 41.9 | 2.74 | **28** | 30.96 | 2.04 | **28** | 28.33 | 0.47 | **28** | 28.1 | 0.4 |
| R5 | 13 | 20.73 | 2.53 | **12** | 13.23 | 0.81 | **12** | 12.46 | 0.62 | **12** | 12.2 | 0.4 |

Table 10: Performance of *Genitor-k*, where *k* represents the fixed number of selected positions for Syswerda's position crossover, in terms of the best and mean number of conflicts. Statistics are taken over 30 independent runs, with 8000 evaluations per run. Min numbers in boldface indicate best known values.

For our study, we run *Genitor-k* with k=10, 50, 100, 150, 200, 250, 300 and 350. We allowed 8000 evaluations per run and performed 30 runs for each problem. The results are summarized in Tables 10 and 11 for minimizing the number of conflicts and in Tables 12 and 13 for minimizing the sum of overlaps. Note that for A6 and A7 there are 299 and 297 requests to schedule respectively. Therefore *Genitor-k* with $k = 300$ and $k = 350$ cannot be run for these two problems. Also note that for example, $k = 200$ does not mean that there are 200 differences in the selected positions between the two parents. The offspring is likely to be very similar to its parents, regardless of the value of *k*, when the parents are similar.

When minimizing the number of conflicts, the worst results are produced by $k = 10$. For $k = 50$, the results improve, best knowns are found for all of the *A* problems; however, for R1, R2, and R3, the best knowns are not found. Starting with $k = 100$ up to $k = 250$ *Genitor-k* finds best known solutions for all the problems. The means and standard deviations are also very similar for all these *k* values; the smallest means and standard deviations correspond to $k = 200$ for the *A* problems and $k = 250$ for the *R* problems (with the exception of R3 for which $k = 200$ produces better results). For $k = 300$, the best knowns are not found anymore for the *A* problems; 300 is very close to the size of the five *A* problems for which it is feasible to run *Genitor-300*. The decay in performance is not as significant for the *R* problems: there is an increase in the means and standard deviations for $k = 300$ and $k = 350$; however, best knowns are still found for four out of the five problems. Note that $k = 400$ would have been a lot closer to the total number of requests for the *R* problems; we believe the performance would have degraded more for the *R* problems for larger *k* values. When minimizing overlaps, we observe trends that are very similar to the ones for minimizing the number of conflicts. $k = 10$ produces poor results, followed by $k = 50$. Similar results are produced by $k = 100, 150, 200, 250$. $k = 150$ results in the smallest means and standard deviations for the *A* problems, while $k = 200$ and $k = 250$ produce best results for the *R* problems. For $k = 300$ and $k = 350$, similarly to minimizing the number of conflicts, the means and standard deviations increase and so do the best solutions found; best knowns are only found for R5.

| Day | Genitor-200 | | | Genitor-250 | | | Genitor-300 | | | Genitor-350 | | |
|-----|-----|------|-------|-----|-------|-------|-----|-------|-------|-----|-------|-------|
| | Min | Mean | Stdev | Min | Mean | Stdev | Min | Mean | Stdev | Min | Mean | Stdev |
| A1 | **8** | 8.56 | 0.56 | **8** | 8.9 | 0.3 | 9 | 11.8 | 1.66 | - | - | - |
| A2 | **4** | 4.0 | 0.0 | **4** | 4.03 | 0.18 | 9 | 13.66 | 1.76 | - | - | - |
| A3 | **3** | 3.0 | 0.0 | **3** | 3.06 | 0.25 | 4 | 9.2 | 2.1 | - | - | - |
| A4 | **2** | 2.0 | 0.0 | **2** | 3.13 | 0.81 | 4 | 8.56 | 1.94 | - | - | - |
| A5 | **4** | 4.3 | 0.46 | **4** | 4.73 | 0.58 | 10 | 13.86 | 2.14 | - | - | - |
| A6 | **6** | 6.06 | 0.25 | **6** | 6.5 | 0.57 | - | - | - | - | - | - |
| A7 | **6** | 6.0 | 0.0 | **6** | 6.06 | 0.25 | - | - | - | - | - | - |
| R1 | **42** | 44.03 | 1.15 | **42** | 44.03 | 0.85 | 43 | 44.26 | 1.01 | 43 | 45.46 | 1.22 |
| R2 | **29** | 29.36 | 0.49 | **29** | 29.4 | 0.49 | **29** | 29.7 | 0.59 | **29** | 30.13 | 0.86 |
| R3 | **17** | 17.33 | 0.4 | **17** | 17.7 | 0.65 | **17** | 17.73 | 0.58 | **17** | 18.63 | 0.8 |
| R4 | **28** | 28.03 | 0.18 | **28** | 28 | 0.0 | **28** | 28.03 | 0.18 | **28** | 28.63 | 0.71 |
| R5 | **12** | 12.1 | 0.3 | **12** | 12.06 | 0.25 | **12** | 12.16 | 0.37 | **12** | 12.6 | 0.81 |

Table 11: Performance of *Genitor-k*, where $k$ represents the fixed number of selected positions for Syswerda's position crossover, in terms of the best and mean number of conflicts. Statistics are collected over 30 independent runs, with 8000 evaluations per run. Min numbers in boldface indicate best known values. The dashes indicate that the permutation solutions for A6 and A7 are shorter than 300 (299 and 297, respectively), and therefore cannot select 300 positions in these permutations.

| Day | Genitor-10 | | | Genitor-50 | | | Genitor-100 | | | Genitor-150 | | |
|-----|------|---------|--------|-----|---------|-------|-----|---------|-------|-----|---------|-------|
| | Min | Mean | Stdev | Min | Mean | Stdev | Min | Mean | Stdev | Min | Mean | Stdev |
| A1 | 149 | 221.53 | 38.85 | 107 | 115.76 | 11.53 | 107 | 107.2 | 0.76 | 107 | 107.1 | 0.54 |
| A2 | 30 | 69.66 | 29.22 | **13** | 15.73 | 3.86 | **13** | 13.43 | 1.54 | **13** | 13.03 | 0.18 |
| A3 | 51 | 122.86 | 36.12 | **28** | 36.26 | 8.19 | **28** | 28.9 | 1.72 | **28** | 28.16 | 0.64 |
| A4 | 59 | 124.5 | 42.25 | **9** | 19.36 | 9.3 | **9** | 9.23 | 0.72 | **9** | 9.06 | 0.36 |
| A5 | 43 | 90.7 | 32.01 | **30** | 33.06 | 3.62 | **30** | 30.36 | 0.96 | **30** | 30.43 | 0.5 |
| A6 | 94 | 145.06 | 33.12 | **45** | 49.6 | 5.54 | **45** | 45.36 | 0.8 | **45** | 45.16 | 0.46 |
| A7 | 67 | 115.66 | 27.96 | **46** | 51.7 | 7.89 | **46** | 46.5 | 2.23 | **46** | 47.63 | 3.9 |
| R1 | 1321 | 1531.13 | 107.35 | 987 | 1139.5 | 76.57 | 914 | 991.13 | 38.19 | 915 | 963.96 | 26.89 |
| R2 | 743 | 961.13 | 81.62 | 557 | 643.86 | 50.0 | 515 | 549.1 | 18.8 | 516 | 540.86 | 15.82 |
| R3 | 480 | 652.5 | 90.37 | 319 | 391.56 | 47.31 | 268 | 305.3 | 20.63 | 269 | 291.3 | 13.36 |
| R4 | 866 | 1069.23 | 74.65 | 768 | 840.23 | 38.79 | 735 | 757.43 | 15.95 | 731 | 752.7 | 14.07 |
| R5 | 208 | 309.03 | 46.3 | **146** | 172.13 | 18.18 | **146** | 151.53 | 7.63 | **146** | 148.23 | 5.26 |

Table 12: Performance of *Genitor-k*, where $k$ represents the fixed number of selected positions for Syswerda's position crossover, in terms of the best and mean sum of overlaps. Statistics are collected over 30 independent runs, with 8000 evaluations per run. Min numbers in boldface indicate best known values.

| Day | Genitor-200 | | | Genitor-250 | | | Genitor-300 | | | Genitor-350 | | |
|-----|-----|------|-------|-----|--------|-------|-----|--------|-------|-----|--------|-------|
| | Min | Mean | Stdev | Min | Mean | Stdev | Min | Mean | Stdev | Min | Mean | Stdev |
| A1 | 107 | 107.1 | 0.74 | 107 | 108.03 | 2.22 | 113 | 157.66 | 26.21 | - | - | - |
| A2 | **13** | 13.2 | 0.92 | **13** | 17.0 | 5.8 | 116 | 185.56 | 33.27 | - | - | - |
| A3 | **28** | 28.9 | 1.6 | **28** | 31.63 | 4.47 | 63 | 106.23 | 26.21 | - | - | - |
| A4 | **9** | 9.1 | 0.4 | **9** | 10.36 | 3.41 | 37 | 78.06 | 25.78 | - | - | - |
| A5 | **30** | 30.6 | 1.3 | **30** | 31.56 | 2.22 | 76 | 160.0 | 36.59 | - | - | - |
| A6 | **45** | 46.33 | 2.7 | **45** | 50.96 | 8.82 | - | - | - | - | - | - |
| A7 | **46** | 47.63 | 4.2 | **46** | 49.93 | 5.39 | - | - | - | - | - | - |
| R1 | 878 | 970.1 | 38.38 | 914 | 968.63 | 31.59 | 935 | 986.7 | 37.9 | 927 | 1008.8 | 42.17 |
| R2 | 512 | 538.43 | 13.94 | 511 | 538.93 | 12.88 | 526 | 551.63 | 12.27 | 532 | 559.46 | 19.7 |
| R3 | 268 | 287.96 | 11.05 | 270 | 292.23 | 12.85 | 272 | 299.43 | 16.3 | 299 | 332.46 | 20.14 |
| R4 | 730 | 752.1 | 12.25 | 734 | 754.53 | 11.89 | 745 | 764 | 13.36 | 743 | 785.36 | 26.63 |
| R5 | **146** | 147.633 | 2.95 | **146** | 147.96 | 3.7 | **146** | 148.6 | 3.84 | **146** | 157 | 10.6 |

Table 13: Performance of *Genitor-k*, where *k* represents the fixed number of selected positions for Syswerda's position crossover, in terms of the best and mean sum of overlaps. Statistics are taken over 30 independent runs, with 8000 evaluations per run. Min numbers in boldface indicate best known values.

In terms of the evolution to the solution, we observe very similar trends for the two objective functions. A typical examples is presented in Figure 5 (minimizing overlaps for R1). *Genitor-k* with $k = 10$ is slower in finding improvements than $k = 50$ which is slower than $k = 100$. $k = 150$ up to $k = 250$ are performing similarly and also similar to the original *Genitor* implementation. $k = 300$ is still moving through the space at a rate that's similar to *Genitor*'s. Only for $k = 350$ does the performance start to decay.

The original implementation of the crossover operator (with a variable number of selected position) was shown to work well not only for our domain but also for other scheduling applications (Syswerda, 1991; Watson, Rana, Whitley, & Howe, 1999; Syswerda & Palmucci, 1991). For our test problems, the results in this subsection show that the number of crossover positions influences the performance of *Genitor*, both in terms of best solutions found and in terms of the rate of finding improvements. For a small number of crossover positions (10 or 50), the solutions found are not competitive, and the improvements are found at a slower rate than in the original *Genitor* implementation. Similarity to *Genitor*'s original performance is obtained for *k* values between 100 and 250. Higher *k* values result in a decay in performance. These results also offer an empirical motivation for the choice of the number of crossover positions in the original *Genitor* implementation. Indeed, in the original implementation, the crossover uses a number of positions randomly selected between one third and two thirds of the total number of requests. This translates for the sizes of problems in our sets to a number of positions that is approximately between 100 and 300.

| Day | Minimizing Conflicts | | | Minimizing Overlaps | | |
|-----|-----|------|-------|-----|------|-------|
|     | Min | Mean | Stdev | Min | Mean | Stdev |
| A1  | **8**   | 8    | 0     | **104** | 104   | 0    |
| A2  | **4**   | 4    | 0     | **13**  | 13    | 0    |
| A3  | **3**   | 3    | 0     | **28**  | 28    | 0    |
| A4  | **2**   | 2    | 0     | **9**   | 9     | 0    |
| A5  | **4**   | 4    | 0     | **30**  | 30    | 0    |
| A6  | **6**   | 6    | 0     | **45**  | 45    | 0    |
| A7  | **6**   | 6    | 0     | **46**  | 46    | 0    |
| R1  | **42***  | 43.4 | 0.7   | 872     | 926.7 | 22.1 |
| R2  | **29**  | 29.9 | 0.3   | 506     | 522.9 | 8.9  |
| R3  | 18  | 18   | 0     | 271     | 283.0 | 6.1  |
| R4  | **28**  | 28.1 | 0.3   | 745     | 765.2 | 10.7 |
| R5  | **12**  | 12   | 0     | **146** | 146   | 0    |

Table 14: Performance of a modified version of SWO where only one request is moved forward for a constant distance 5. For both minimizing conflicts and minimizing the sum of overlaps, the request is randomly chosen. All statistics are collected over 30 independent runs, with 8000 evaluations per run. A ∗ indicates that the best value is better than the corresponding SWO result. Min numbers in boldface indicate best known values.

### 6.5.2 The Effect of Multiple Moves on SWO

We hypothesize that the multiple moves present in SWO are necessary for its performance. To test this hypothesis, we start by investigating the effect of moving forward only one request. This is somewhat similar to the shifting operator present in RLS: a request is shifted forward in the permutation. However, to implement the SWO reprioritization mechanism, we restrict both the chosen request to be moved and the position where it gets moved. For minimizing the conflicts, one of the bumped requests is randomly chosen; for minimizing overlaps, one of the requests contributing to the sum of overlaps is randomly chosen. For both minimizing the conflicts and minimizing the sum of overlaps the chosen request is moved forward for a constant distance of five[10]. We call this new algorithm *SWO1Move*. The results obtained by running *SWO1Move* for 30 runs with 8000 evaluations per run are presented in Table 14. The entries with a ∗ indicate that the value produced by *SWO1Move* is better than the corresponding SWO result. The initial solutions are identical to the solutions produced using the flexibility heuristic for initializing SWO.

When minimizing conflicts, *SWO1Move* performs as well as SWO (in fact, it finds the best known solution for R1 as well). When minimizing the sum of overlaps, the performance of SWO for the *R* problems worsens significantly when only one task is moved forward. Previously, we implemented *SWO1Move* for minimizing overlaps by moving forward the request that contributes most to the total overlap (Barbulescu et al., 2004c). Randomly choosing the request to be moved forward improved the performance of *SWO1Move*. Randomization is useful because SWO can become trapped in cycles (Joslin & Clements, 1999); however, the improvement is not enough to equal the performance of SWO when minimizing overlaps for the new days of data. In fact, longer runs of *SWO1Move* with a random choice of the request to be moved (30 runs with 50,000 evaluations) produce solutions that are still worse than those obtained by SWO. These results support the conjecture that the performance of SWO is due to the simultaneous moves of requests.

---

10. We tried other values; on average, a value of five seems to work best.

| Day | k=10 | | | k=20 | | | k=30 | | | k=40 | | |
|-----|------|------|-------|------|--------|-------|------|--------|-------|------|--------|-------|
| | Min | Mean | Stdev | Min | Mean | Stdev | Min | Mean | Stdev | Min | Mean | Stdev |
| R1 | 840 | 862.5 | 11.28 | 815 | 829.77 | 8.45 | 798 | 820.63 | 8.12 | 825 | 841.13 | 8.02 |
| R2 | 512 | 530.2 | 9.18 | 498 | 506.53 | 5.25 | 493 | 508.97 | 5.26 | 508 | 526.26 | 6.25 |
| R3 | 284 | 291.36 | 4.65 | 266 | 268.9 | 2.21 | 266 | 271.07 | 3.52 | 266 | 273.2 | 3.54 |
| R4 | 764 | 778.57 | 8.45 | 749 | 757.3 | 6.16 | 740 | 744.47 | 2.6 | 737 | 747.2 | 5.06 |

Table 15: Performance of a modified version of SWO where $k$ of the requests contributing to the sum of overlaps are moved forward for a constant distance 5. All statistics are collected over 30 independent runs, with 8000 evaluations per run.

We attribute the discrepancy in the *SWO1Move* performance for the two objective functions to the difference in the discretization of the two search spaces. When minimizing conflicts, *SWO1Move* only needs to identify the requests that cannot be scheduled. More fine tuning is needed when minimizing the sum of overlaps; besides identifying the requests that cannot be scheduled, *SWO1Move* also needs to find the positions for these requests in the permutation such that the sum of overlaps is minimized. We conjecture that this fine tuning is achieved by simultaneously moving forward multiple requests.

Next, we investigate the changes in performance when an increasing number of requests (from the requests contributing to the objective function) are moved forward. We design an experiment where a constant number of the requests involved in conflicts are moved forward. To do this, we need to decide how many requests to move and which ones. Moving two or three requests forward results in small improvements over the results in Table 14. Therefore, we run multiple versions of SWO moving $k$ requests forward, with k = 10, 20, 30, 40. We determined empirically that when moving multiple requests (more than five) forward, choosing them at random as opposed to based on their contribution to the sum of overlaps hurts algorithm performance. To determine which requests are moved forward, at each step we sort the requests contributing to the sum of overlaps in decreasing order of their contribution and only move forward the first $k$ (or all of them, if $k$ is greater than the number of requests contributing to the sum of overlaps).

The results obtained for $R1$, $R2$, $R3$ and $R4$ are summarized in Table 15. For the $A$ problems, all these new SWO versions find best known solutions. We did not include $R5$ in this study because the SWO greedy initial permutation computed for $R5$ corresponds to a best known value schedule. The results show a general performance improvement as $k$ grows from 10 to 20. $k = 20$ and $k = 30$ produce similar performance for $R1$, $R2$ and $R3$. For $R4$, $k = 30$ results in better performance than $k = 20$. $k = 40$ results in worsening performance for $R1$ and $R2$. Note that the algorithm performance for $R3$ does not change when $k >= 20$. This is not surprising; since good solutions (in terms of overlaps) for this problem correspond to schedules with a small number of overlapping tasks, moving forward 20 requests or more means moving most of the requests in conflict once good solutions are found. The results indicate that for the problems in our set, when minimizing overlaps, if SWO is allowed to only move forward a constant number $k$ of requests, $k = 30$ seems to be a good choice.

The results in this section support the hypothesis that moving multiple requests forward is necessary to obtain good SWO performance. First, we showed that moving only one request forward (or a small number of requests, smaller than 30 for the $R$ problems) results in inferior SWO performance. Second, as the number of requests moved forward is increased (from 10 up), the performance of SWO improves.

## 7. New Algorithm: Attenuated Leap Local Search

The empirical data and analyses suggest that the key to competitive performance on this application is moving as quickly as possible across the plateaus. Two of the competitive algorithms, Genitor and SWO, perform multiple moves. A simpler algorithm, RLS, actually finds more best known solutions in 8000 evaluations, even though it does not perform multiple moves. RLS does however, perform a significant number of "neutral moves" between solutions with the same evaluation. Given this, we conjecture that a version of local search that performs multiple moves before evaluating the result may be even better suited to this application. The intuition behind this conjecture is that the search should sample at greater distances (i.e., longer than a single move) to more quickly find exits from plateaus.

We modified the RLS move operator as follows: choose a number of pairs of positions and apply shifting to these pairs, one after another, without building the schedule after each shift; build the schedule only after shifting has been applied for the designated number of pairs. In our first version, we tried a static number of shifts (10 turned out to be the best value); however, it performed no better and sometimes worse than the original move operator. We next conjectured that as search progresses to better solutions, the number of shifts should also decrease because the probability of finding detrimental moves (rather than improving) increases significantly as well. The better the solution, the fewer exits are expected and the harder they are to find.

We implemented a multiple move hill-climber with a variable move count operator: given a decay rate, we start by shifting ten requests, then nine, eight etc. We chose to decrement the number of shifts for every 800 evaluations; we call this version of hill-climbing *Attenuated Leap Local Search* (ALLS). This is similar to the idea behind the "temperature dependent" hill-climbing move operator implemented by Globus et al. (2004), for which the number of requests to move is chosen by random but biased such that a large number of requests are moved early in the search while later only a few requests are moved[11]. Hill-climbing with the temperature dependent operator produced better results for EOS than simply choosing a random number of requests to move.

ALLS performs remarkably well. As shown in Table 16, it finds best known values for all problems using conflicts and all but two of the problems using overlaps (as does

---

11. The operator is similar to the temperature dependent behavior in simulated annealing; this explains the name of the operator.

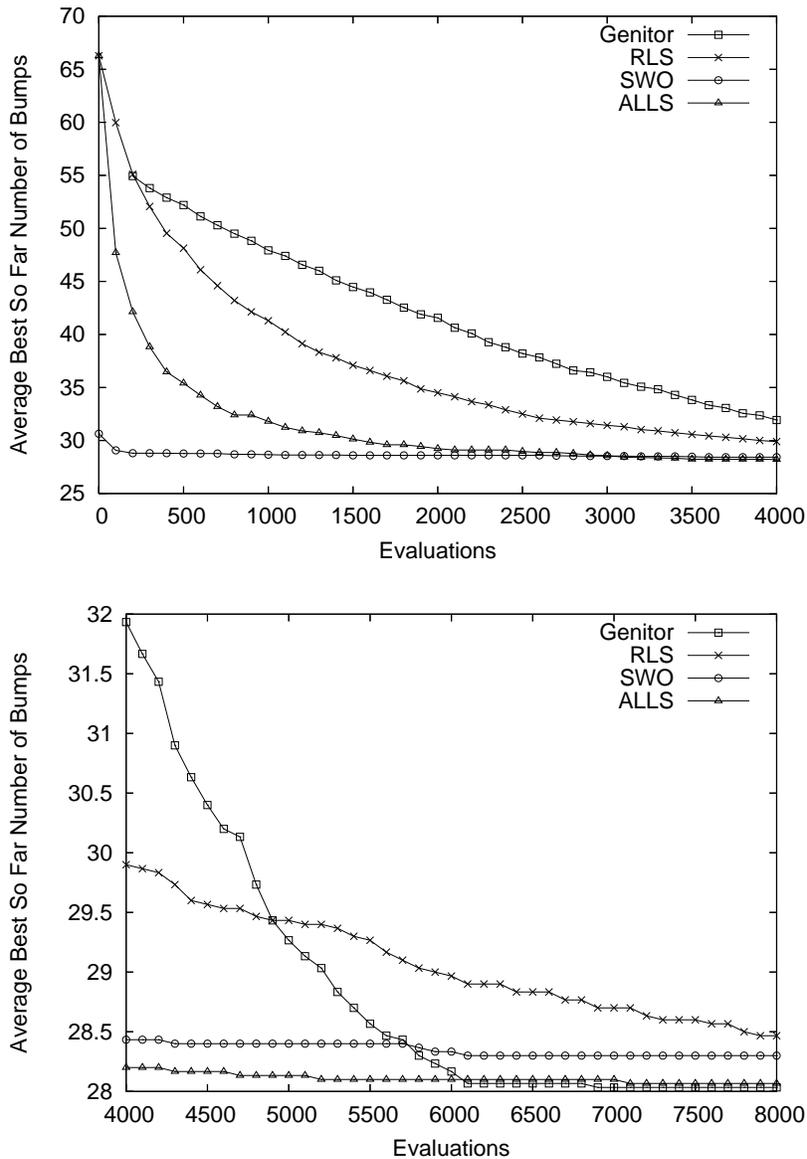

Figure 6: Evolutions of the average best value obtained by *Genitor*, RLS, SWO and *ALLS* during 8000 evaluations, over 30 runs. The improvement over the first 4000 evaluations is shown in the top figure. The last 4000 evaluations are depicted in the bottom figure; note that the scale is different on the y-axis. The graphs were obtained for $R4$; best solution value is 28.

RLS). Additionally, it finds better best values than all the algorithms in our set for the two problems with non-best solutions. In fact, a single tailed, two sample t-test comparing ALLS to RLS shows that ALLS finds statistically significantly better solutions ($p < 0.023$) on both conflicts and overlaps for the five more recent days.

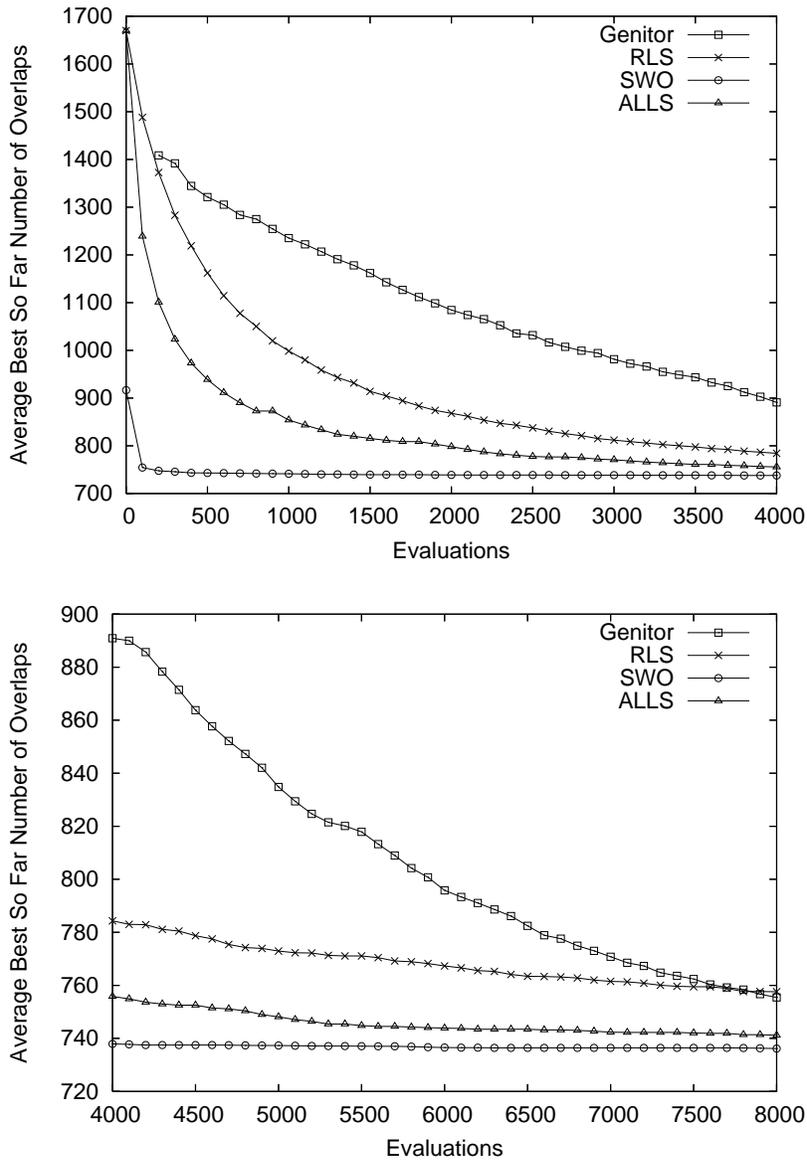

Figure 7: Evolutions of the average best value obtained by *Genitor*, RLS, SWO and *ALLS* during 8000 evaluations, over 30 runs. The improvement over the first 4000 evaluations is shown in the top figure. The last 4000 evaluations are depicted in the bottom figure; note that the scale is different on the y-axis.The graphs were obtained for $R4$; best solution value is 725.

In Section 5, we discussed a comparison across all algorithms (again at $p < 0.005$). Under this much more restrictive performance comparison, ALLS still outperforms RLS, SWO and *Genitor* for most of the pair-wise tests. Both when minimizing conflicts and when minimizing overlaps, ALLS significantly outperforms all other algorithms on R1. When minimizing conflicts, ALLS outperforms for all but five of the twelve pair-wise tests on the other four days (for which the difference was not significant). The exceptions are: R2, R3, R4, and R5 for *Genitor* and R4 for RLS. When minimizing overlaps, ALLS significantly outperforms *Genitor* for R2, RLS for R3, *Genitor* for R4 and SWO for R5; the rest of the pair-wise comparisons were not statistically significant at $p < 0.005$. It is clear that ALLS is at least as good as the best algorithms and outperforms them on most days of data.

| Day | Minimizing Conflicts | | | Minimizing Overlaps | | |
|-----|-----|------|-------|-----|--------|--------|
|     | Min | Mean | Stdev | Min | Mean   | Stdev  |
| A1  | **8**  | 8.2   | 0.4  | **104** | 107.1  | 1.24  |
| A2  | **4**  | 4.0   | 0.0  | **13**  | 13.0   | 0.0   |
| A3  | **3**  | 3.0   | 0.0  | **28**  | 28.33  | 1.3   |
| A4  | **2**  | 2.03  | 0.18 | **9**   | 9.13   | 0.73  |
| A5  | **4**  | 4.1   | 0.3  | **30**  | 30.23  | 0.43  |
| A6  | **6**  | 6.0   | 0.0  | **45**  | 45.0   | 0.0   |
| A7  | **6**  | 6.0   | 0.0  | **46**  | 46.0   | 0.0   |
| R1  | **42** | 42.63 | 0.72 | 785     | 817.83 | 27.07 |
| R2  | **29** | 29.1  | 0.3  | 490     | 510.37 | 19.14 |
| R3  | **17** | 17.5  | 0.57 | **250** | 273.33 | 43.68 |
| R4  | **28** | 28.07 | 0.25 | **725** | 740.07 | 19.56 |
| R5  | **12** | 12.0  | 0.0  | **146** | 146.03 | 0.19  |

Table 16: Statistics for the results obtained in 30 runs of *ALLS*, with 8,000 evaluations per run. The best and mean values as well as the standard deviations are shown. Bold indicates best known values.

ALLS also finds improving solutions faster than both *Genitor* and RLS (see Figures 6 and 7 for R4 on both conflicts and overlaps). ALLS achieves such good performance by combining the power of finding good solutions fast using multiple moves in the beginning of the search with the accuracy of locating the best solutions using one-move shifting at the end of the search.

In 6.4 we showed that as the solutions improve the random walks on plateaus become longer. Two hypotheses support this observation: 1) The plateaus are bigger 2) The plateaus are harder to escape because there are fewer exits. These two hypotheses are consistent if the missing exits are replaced by moves of equal value. They are not consistent if the exits are replaced by worse moves. Our ALLS design implicitly assumes the latter. If the exits were replaced by equal moves then as the search progresses more moves would be needed per each large step[12]. In fact, we ran some tests where we increased the number of moves as search progresses and we found that this can significantly worsen the performance. For example, for $R1$ when minimizing overlaps, shifting initially ten requests and then increasing the number of shifted requests by 1 every 800 iterations (instead of decreasing it as in ALLS) results in a minimum overlap of 885, with a mean of 957.97 and a standard deviation of 51.36, which is significantly worse than the corresponding ALLS result.

---

12. We wish to thank the anonymous reviewer for this insightful observation.

## 8. Conclusion

A key algorithm characteristic for AFSCN appears to be multiple moves. In fact, this observation might hold for other oversubscribed scheduling problems as well. Globus et al. (Globus et al., 2004) found that when solving the oversubscribed problem of scheduling fleets of EOS using hill-climbing, moving only one request at a time was inefficient. Their temperature dependent hill-climbing operator proved to work better than simply choosing a random number of requests to move. As in our domain, a permutation representation and a greedy deterministic schedule builder are used. We conjecture that their schedule builder also results in multiple permutations being mapped to the same schedule, and therefore that plateaus are present in the EOS search space as well. The fact that moving more than one request improved the results suggests that our conjecture could also hold for EOS scheduling: multiple moves might speed up plateau traversal for this domain as well.

We developed and tested four hypotheses explaining the performance of three competitive algorithms for a real scheduling application. We found that all of the hypotheses held to varying degrees. Based on the evidence, we designed a new algorithm that combined what appeared to be critical elements of the best performing algorithms and produced an algorithm that performed better than the original ones. Our results suggest that multiple moves are a useful algorithm feature to obtain good performance results for AFSCN scheduling. Alternatively, it is possible that in fact only one move during each iteration would be enough to obtain good performance, but it is difficult to identify which request to move. Future research in this direction will examine heuristics such as combining HBSS and SWO to decide which request to move forward, as well as heuristics to find where to move the request to guarantee a change in the schedule. Also as future research, we will be testing other oversubscribed scheduling applications to determine to what extent our analyses and results generalize: do they exhibit the same characteristics and are they amenable to the same kind of solution?

## Acknowledgments

This research was supported in part by a grant from the Air Force Office of Scientific Research, Air Force Materiel Command, USAF under grant number F49620-03-1-0233. Adele Howe was also supported by the National Science Foundation under Grant No. IIS-0138690. Any opinions, findings, and conclusions or recommendations expressed in this material are those of the author(s) and do not necessarily reflect the views of the National Science Foundation. The U.S. Government is authorized to reproduce and distribute reprints for Governmental purposes notwithstanding any copyright notation thereon.